\documentclass{article} % For LaTeX2e
\usepackage{iclr2027_conference,times}

\usepackage{amsmath,amsfonts,bm}

\def\eqref#1{equation~\ref{#1}}
\def\1{\bm{1}}

\DeclareMathAlphabet{\mathsfit}{\encodingdefault}{\sfdefault}{m}{sl}
\SetMathAlphabet{\mathsfit}{bold}{\encodingdefault}{\sfdefault}{bx}{n}

\usepackage{hyperref}
\usepackage{url}
\usepackage{graphicx}

\usepackage{booktabs}       % professional-quality tables
\usepackage{amsfonts}       % blackboard math symbols
\usepackage{nicefrac}       % compact symbols for 1/2, etc.
\usepackage{microtype}      % microtypography
\usepackage{xcolor}         % colors
\usepackage{float}
\usepackage{multirow}
\usepackage{colortbl}
\usepackage{utfsym}
\usepackage{tcolorbox}
\usepackage{alltt}
\usepackage{enumitem}
\usepackage{hyperref}
\usepackage{amsmath}
\usepackage{xcolor}
\usepackage{makecell}
\usepackage{tabularx}
\usepackage{array}
\usepackage{placeins}
\usepackage{listings}
\usepackage{url}
\usepackage{caption}
\usepackage{wrapfig}
\usepackage{soul}
\definecolor{headerblue}{RGB}{240, 248, 255}
\sethlcolor{headerblue}
\newcommand{\myfinding}[1]{%
  \vspace{0.3em}\noindent\hl{\textbf{#1.}}}

\definecolor{mypink}{rgb}{.99,.91,.95}
\newcolumntype{Y}{>{\centering\arraybackslash}X}
\definecolor{NiceGreen}{RGB}{2,158,115} 
\definecolor{NiceRed}{RGB}{213,94,0}
\newcommand{\up}[1]{\textcolor{NiceGreen}{\scriptsize$\uparrow$#1}}
\newcommand{\down}[1]{\textcolor{NiceRed}{\scriptsize$\downarrow$#1}}
\newcommand{\Ev}[1]{\hyperref[ev:#1]{[A#1]}}
\usetikzlibrary{shadows}
\definecolor{kleinblue}{RGB}{0, 47, 167} 
\definecolor{kleinblue2}{RGB}{20, 20, 125} 
\definecolor{green}{RGB}{146, 192, 113} 
\newcommand*\circled[1]{%
\tikz[baseline=(char.base)]{
  \node[shape=circle, fill=kleinblue2, draw=white, thick, drop shadow={shadow xshift=0.2ex,shadow yshift=-0.2ex}, inner sep=1pt] (char) {\textcolor{white}{#1}};
}}

\setlist[itemize]{leftmargin=1.5em, labelindent=0.5em}
\AtBeginDocument{%
  \renewcommand{\lstlistingname}{Table}%
}

\title{Parse Before You Execute: Efficient Spatial Reasoning with Executable Primitives}

\author{Xiwei Liu$^{1}$, Yulong Li$^{1}$, Xinling Zhuang$^{2}$, Xuhui Li$^{1}$, Yichen Li$^{3}$, Haolin Yang$^{1}$, \\
\textbf{Zhixiang Lu$^{4}$, Yutong Xie$^{1}$\thanks{Corresponding author.}} \\
{\small $^{1}$Mohamed bin Zayed University of Artificial Intelligence \quad
$^{2}$The Chinese University of Hong Kong} \\
{\small $^{3}$Huazhong University of Science and Technology \quad
$^{4}$University of Liverpool}
}

\iclrfinalcopy % Uncomment for camera-ready version, but NOT for submission.
\begin{document}

% \vspace{-1em}
\maketitle
\vspace{-1em}
\begin{abstract}
% Spatial reasoning in Vision-Language Models (VLMs) has increasingly benefited from tool-integrated pipelines that ground computation in explicit geometry. 
% However, existing systems still compile queries into free-form artifacts, such as code, task constraints, tool plans, or gathered evidence, all of which remain in open output spaces. As a result, they can often produce plausible reasoning traces even when the query exceeds their actual representational reach.
% We identify this failure mode as under-committed query compilation. To address it, we propose \textbf{PEPS}, a training-free agentic paradigm \underline{\textbf{P}}arsing \underline{\textbf{E}}xecutable \underline{\textbf{P}}rimitives for \underline{\textbf{S}}patial reasoning under a typed FESM primitive schema over Frame, Entity, State, and Metric. The Parser agent retrieves a closed requirement set from a primitive library; an Executor agent grounds the required primitives through fixed geometric APIs and deterministic computation; and a Verifier agent checks whether the resulting answer is fully supported by the query, visual evidence, and execution trace, returning typed feedback that distinguishes missing primitives from genuine schema gaps. Extensive experiments on challenging spatial reasoning benchmarks show that PEPS improves over strong tool-integrated baselines, transfers its evolved primitive schema across benchmarks, and exposes compilation success as an interpretable predictor of downstream correctness. Source code is available at \url{https://anonymous.4open.science/r/PEPS-A31A}.

Spatial reasoning in Vision-Language Models (VLMs) has increasingly benefited from tool-integrated pipelines that ground computation in explicit geometry. However, existing systems often compile queries into relatively open-ended artifacts, such as code, task constraints, tool plans, or gathered evidence, leaving what must be resolved before execution under-specified. This can lead to unnecessary tool acquisition and iterative refinement, while still producing plausible reasoning traces from incomplete spatial commitments. We identify this failure mode as under-committed query compilation. To address it, we propose \textbf{PEPS}, a training-free agentic paradigm \underline{\textbf{P}}arsing \underline{\textbf{E}}xecutable \underline{\textbf{P}}rimitives for \underline{\textbf{S}}patial reasoning under a typed FESM schema over Frame, Entity, State, and Metric. PEPS follows a \textit{parse-before-execute} principle: the Parser commits each query to a compact set of executable requirements; the Executor selectively grounds the required primitives through fixed geometric APIs and deterministic computation; and the Verifier checks execution sufficiency, enabling early termination or targeted repair. Extensive experiments across challenging spatial reasoning benchmarks show that PEPS consistently improves reasoning accuracy while substantially reducing refinement rounds, API cost, and end-to-end latency over strong tool-integrated baselines. These results demonstrate that explicit requirement commitment can make tool-integrated spatial reasoning not only more reliable, but also more execution-efficient. Source code is available at \url{https://github.com/Xiwei-web/PEPS}.
\end{abstract}

\section{Introduction}
\label{sec:introduction}

Spatial reasoning is rarely a matter of exhaustively describing everything in a scene. Cognitive studies suggest that humans reason about space in a question-driven way, selecting the scene structure that is relevant to the task at hand~\citep{tversky1991spatial, byrne1989spatial}. Someone moving a sofa through a doorway, for instance, needs the doorway width, the sofa's largest dimension, and whether it can be tilted, while many other visual details are irrelevant. This observation suggests a useful design principle for spatial reasoning systems: \textit{\textcolor{kleinblue2}{before perception tools are invoked or geometric computation begins, a spatial reasoning system should first commit to what the query actually requires it to know.}}

Recent spatial reasoning systems have made substantial progress by moving part of the reasoning process out of unconstrained VLM inference and into explicit tools~\cite{suris2023vipergpt, gupta2023visual, jia2025omnispatial}. Prior work has shown the value of offloading geometric computation to deterministic modules~\cite{luo2026pyspatial, chen2025geometrically}, expanding or adapting tool use through APIs~\cite{marsili2025visual, wu2025transductive}, and curating sufficient evidence for downstream reasoning~\cite{guo2025pursuing}. Together, these advances suggest that reliable spatial reasoning depends not only on stronger models, but also on how perception, computation, and evidence acquisition are organized. Yet existing systems determine what is needed through open-ended plans, programs, formalizations, or iterative evidence requests. When the required information is not committed before execution, the agent may acquire unnecessary evidence, invoke redundant tools, or repeatedly refine an initially incomplete reasoning trajectory.

\begin{figure}[t!] % [htbp]
    \centering
    \includegraphics[width=0.9\textwidth]{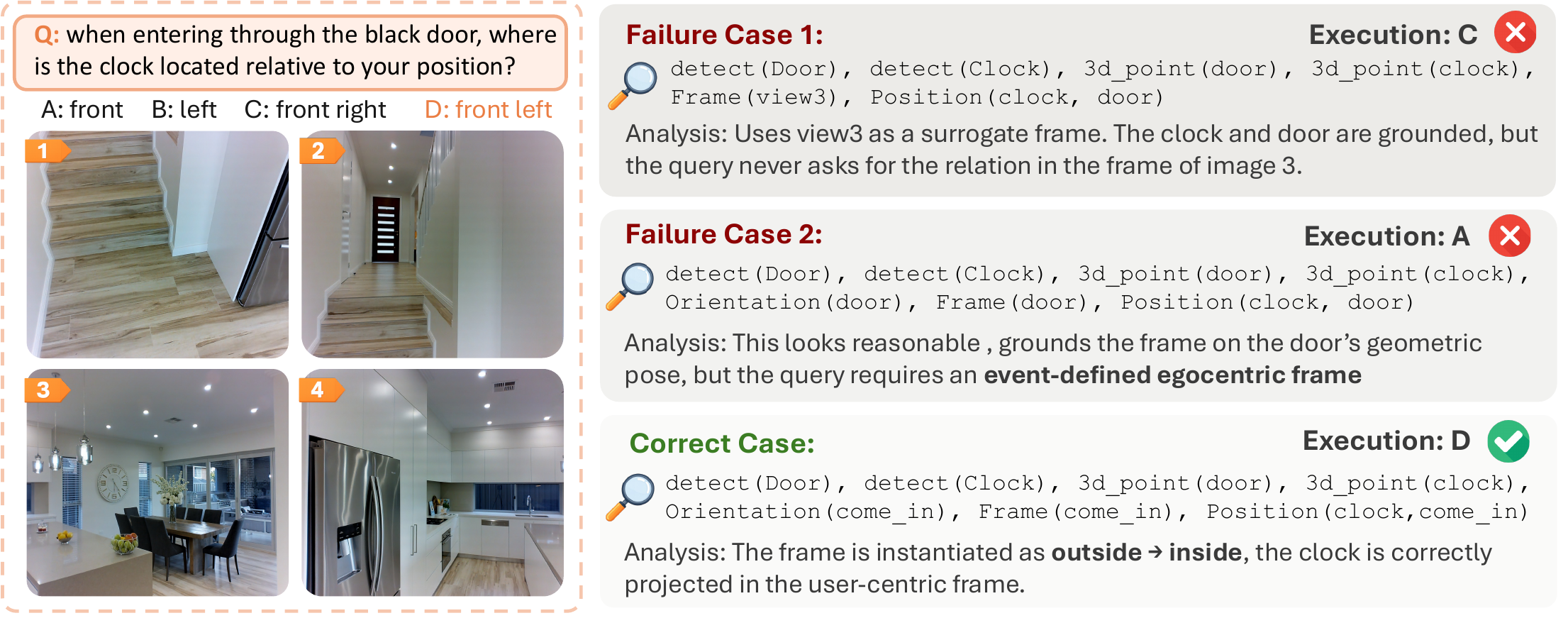} 
    \vspace{-0.8em}
    \caption{Plausible but incomplete requirement commitments can mislead spatial reasoning.  } 
    \label{fig:motivation case}
    \vspace{-3em}
\end{figure}

The missing piece is therefore not only another tool, but an explicit interface between the query and its execution. In existing spatial reasoning systems, the artifact compiled from a query, whether free-form code, a natural-language formalization, a tool-selection plan, or an information set, often remains in an open output space~\cite{marsili2025visual, luo2026pyspatial, chen2025geometrically}. Such artifacts may specify how to proceed without explicitly committing to what must be resolved before execution. As a result, a system can produce a plausible execution trace even when the required spatial commitments are wrong or incomplete. Figure~\ref{fig:motivation case} illustrates this failure mode: the tools are sufficient and the geometry is correct, yet execution fails because the query is compiled into the wrong requirements. We call this \textit{under-committed query compilation}. Beyond correctness, such incomplete commitments also create computational redundancy: missing requirements must be discovered downstream through additional acquisition, replanning, or refinement.

We therefore follow a simple principle: \textit{parse before you execute}. Rather than asking a VLM to freely generate what to do next, we first compile the query into a compact set of executable spatial primitives retrieved from a predefined library. Each primitive specifies one requirement that must be resolved before deterministic computation, such as which reference frame should be used, which entities must be grounded, which geometric states must be acquired, and which measurable relation should be computed. We organize these requirements into a four-part schema, \textbf{FESM}: Frame, Entity, State, and Metric. This typed interface makes the execution boundary explicit: only required primitives need to be instantiated, while unresolved requirements can be identified and repaired without restarting an open-ended reasoning process. Parsing therefore serves not only as an inspectable commitment, but also as a mechanism for selective execution and early termination.

We instantiate this principle in \textbf{PEPS}, a multi-agent framework for \underline{\textbf{P}}arsing \underline{\textbf{E}}xecutable \underline{\textbf{P}}rimitives for \underline{\textbf{S}}patial Reasoning. The Parser commits each query to a compact FESM requirement set. The Executor selectively grounds the required primitives through geometric APIs and deterministic computation, reusing available results whenever possible. The Verifier checks whether the resulting answer is fully supported by the query, visual evidence, and execution trace; sufficiently supported traces terminate immediately, while rejected traces receive typed feedback that distinguishes missing-slot failures from genuine primitive gaps. This compile-execute-verify design targets both reasoning reliability and execution efficiency by avoiding unnecessary acquisition and repeated refinement. We evaluate PEPS on MMSI-Bench~\cite{yang2025mmsi}, ViewSpatial-Bench~\cite{li2025viewspatial}, and OMNI3D-Bench~\cite{marsili2025visual}. PEPS consistently improves spatial reasoning performance across different VLM backbones and, compared with the strongest iterative tool-integrated baseline, reduces API cost by 15.0-23.5\% and mean end-to-end latency by 32.3-38.2\%, with 83.0-93.0\% of queries terminating after a single round. Our contributions are threefold:

% \vspace{-0.8em}
\noindent\circled{1} We characterize under-committed query compilation as a failure mode in tool-integrated spatial reasoning, where plausible intermediate artifacts may leave query-required spatial commitments unresolved. This motivates making query-to-requirement commitment explicit before execution.

% % \vspace{-0.8em}
\noindent\circled{2} We introduce PEPS, a \textit{parse-before-execute} framework, and FESM, a typed interface over Frame, Entity, State, and Metric. Compact executable primitives enable selective tool execution, deterministic computation, typed failure repair, and early termination rather than open-ended iterative acquisition.

% \vspace{-0.8em}
\noindent\circled{3} Extensive experiments show that explicit requirement commitment yields consistent gains across benchmarks and VLM backbones. Beyond accuracy improvements, PEPS reduces API cost, refinement rounds, and end-to-end latency over iterative tool-integrated baselines. Further analyses show that primitive parsing also benefits existing agentic pipelines as a committed-compilation front end.

\section{Methodology}
\label{sec:method}

\begin{figure}[t!] % [htbp]
    \centering
    \includegraphics[width=1.0\textwidth]{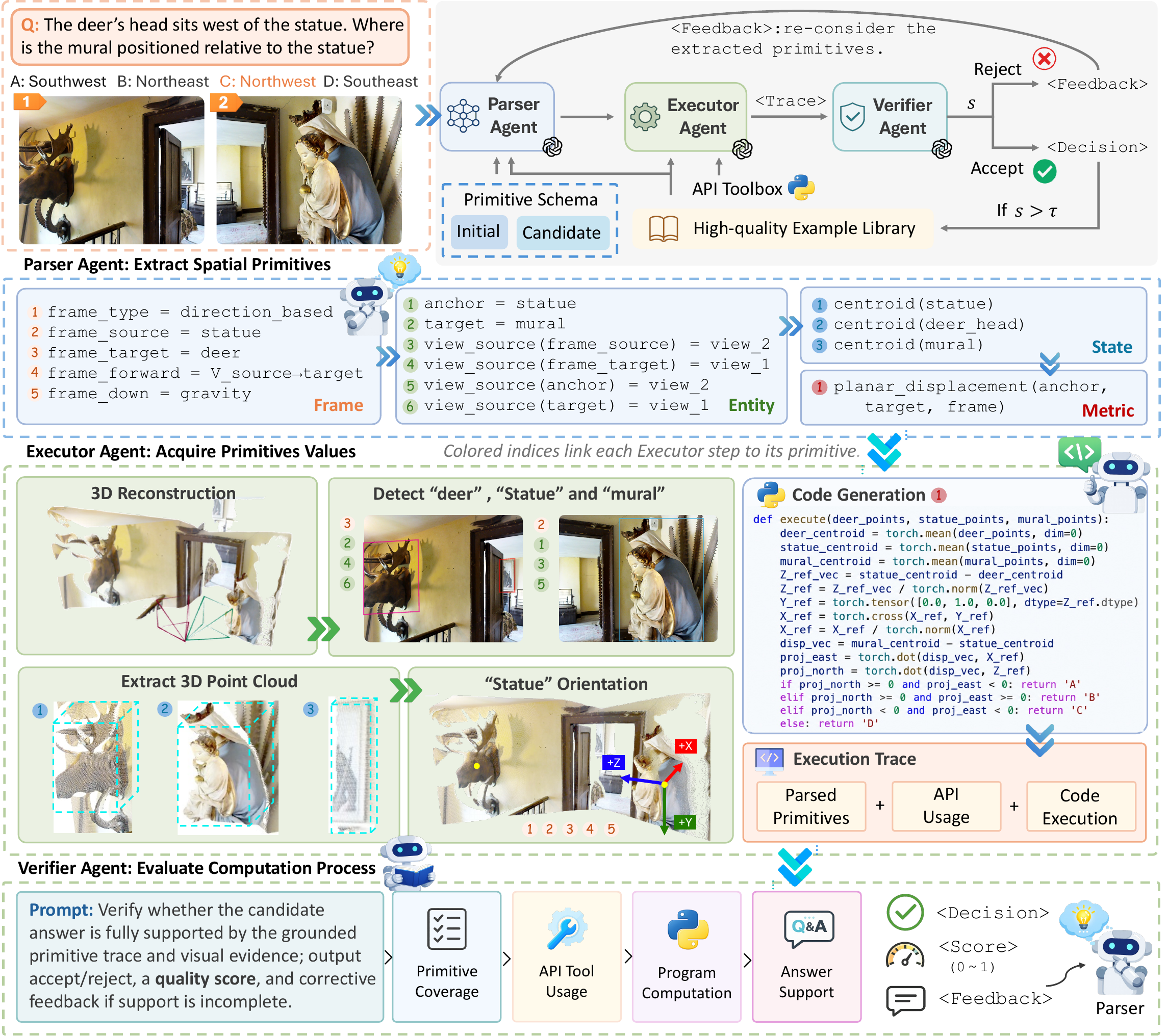} 
    \vspace{-1.4em}
    \caption{\textbf{Overall Paradigm.} The Parser retrieves Frame, Entity, State, and Metric primitives from the query. The Executor grounds them with a fixed geometric toolbox and deterministic code, producing a candidate answer and trace. The Verifier outputs \texttt{<Decision>}, \texttt{<Score>}, and \texttt{<Feedback>}; rejected candidates are refined, and high-scoring accepted ones are stored for reference.} 
    \label{fig:overview}
    \vspace{-1em}
\end{figure}

\paragraph{Problem Formalization.}
We consider language-conditioned spatial reasoning over a scene observed from one or multiple views.
Let $\mathcal{I}=\{I_1,\ldots,I_M\}$ denote the input images, $q$ a natural-language spatial query, and $a^*$ the ground-truth answer.
Given a typed primitive library $\mathcal{L}$ over Frame, Entity, State, and Metric, our goal is to recover a compact sufficient executable primitive set $\mathcal{P} \subseteq \mathcal{L}$ that specifies the requirements needed to answer $q$ through deterministic geometric computation.

PEPS realizes this objective through a closed loop of parsing, execution, and verification.
The Parser retrieves a candidate requirement set $\mathcal{P}\subseteq\mathcal{L}$, the Executor selectively instantiates the unresolved primitives through a fixed API toolbox $\mathcal{S}$, and the Verifier evaluates whether the resulting candidate is sufficiently supported by the query, visual evidence, and execution trace:
\begin{equation}
\mathcal{P}=\mathcal{A}_{\mathrm{parse}}(\mathcal{I},q,\mathcal{L}), \qquad
(\hat{a},T)=\mathcal{A}_{\mathrm{exec}}(\mathcal{I},q,\mathcal{P},\mathcal{S}), \qquad
(v,s,F)=\mathcal{A}_{\mathrm{verify}}(\mathcal{I},q,\hat{a},T),
\end{equation}
where $T$ is the execution trace, $v \in \{\texttt{accept},\texttt{reject}\}$ is the verification decision, $s \in \mathbb{R}$ is the quality score, and $F$ is the structured feedback returned when $v=\texttt{reject}$. The loop terminates once a candidate is accepted as the final answer or reaches the maximum loop iteration limitation.

\subsection{Spatial Primitives Schema Construction}
\begin{wrapfigure}{r}{0.6\linewidth}
\centering
\vspace{-1em}
\captionsetup{skip=2pt}
\includegraphics[width=\linewidth]{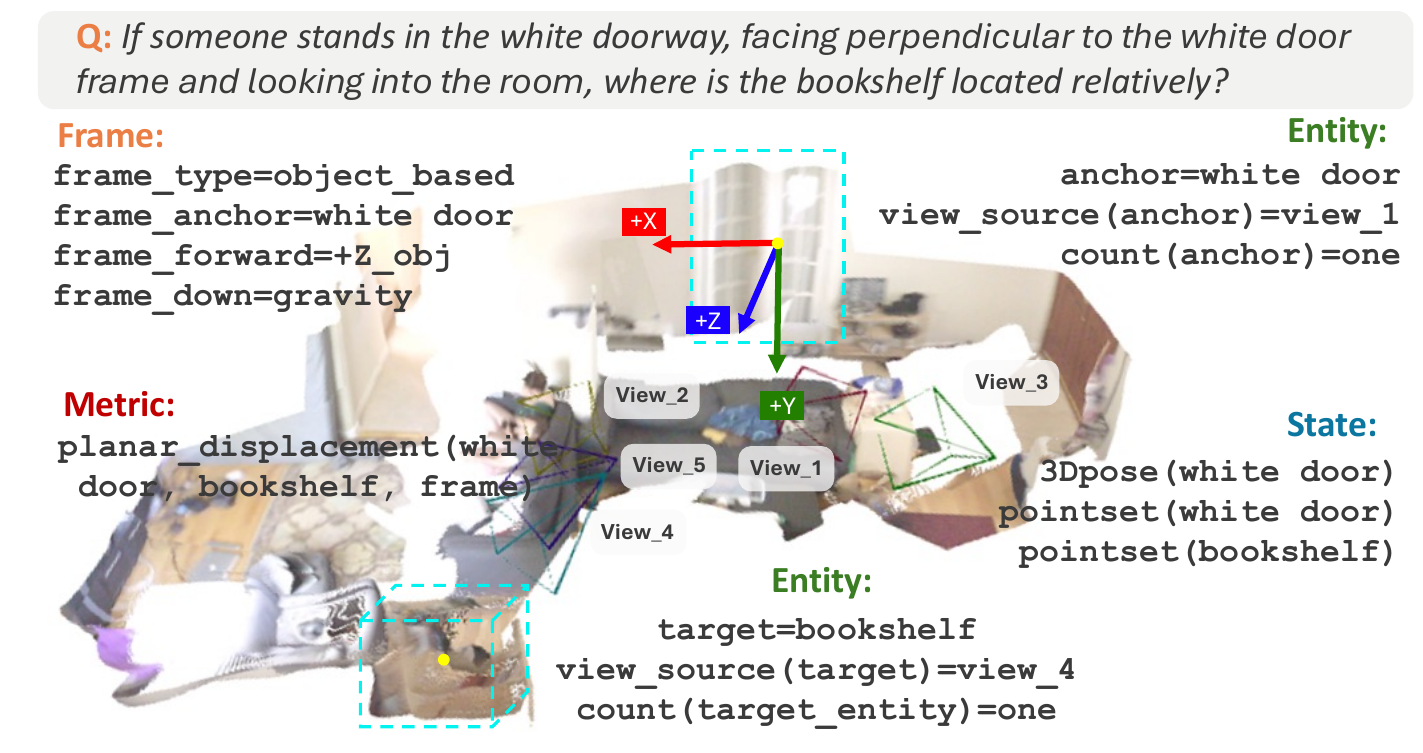}
\caption{FESM-dimensional decomposition of a query.}
\label{fig:schema}
\end{wrapfigure}

\textcolor{kleinblue2}{\textit{What must be explicitly resolved before a spatial query can be executed deterministically?}}
To answer this, we construct a typed FESM schema as an executable interface between query interpretation and tool execution (Refer Appendix~\ref{app:schema}).
FESM is not a taxonomy of surface spatial relations, but follows their execution dependencies as illustrated in Figure~\ref{fig:schema}.
Frame primitives determine the reference coordinate system, Entity primitives identify what to ground, State primitives specify the geometric variables to acquire, and Metric primitives define the relation to compute.
Together, they specify what must be instantiated while avoiding unnecessary acquisition.
We induce the initial FESM schema with a two-stage prompting pipeline over questions sampled from MMSI-Bench~\cite{yang2025mmsi}, ViewSpatial-Bench~\cite{li2025viewspatial}, and OMNI3D-Bench~\cite{marsili2025visual}.
At the per-question level, GPT-4o~\cite{openai_gpt4o_2024} decomposes each query into candidate FESM requirements without solving the question or proposing tool calls.
These decompositions are then aggregated by merging equivalent requirements, parameterizing object-specific variables, removing candidates unsupported by the tool suite, and recording supporting question IDs.
Detailed prompts are provided in Appendix Tables~\ref{tab:app_prompt_per_question_decomposition} and~\ref{tab:app_prompt_schema_aggregation}.
We further evaluate schema transfer to MindCube-Tiny~\cite{yin2025spatial} and OmniSpatial~\cite{jia2025omnispatial} in Section~\ref{sec:ablation}.

\subsection{Parser Agent}
\label{subsec:parser}
The Parser Agent is responsible for committed compilation. Given the input images $\mathbf{I}$, query $q$, and current library $\mathcal{L}$, it retrieves a structured requirement set $\mathcal{P} = \mathcal{A}_{\mathrm{parse}}(\mathbf{I}, q, \mathcal{L})$ that specifies what must be grounded before deterministic spatial computation. Its output is restricted to primitive instances in $\mathcal{L}$, making the parsed result an explicit and inspectable statement of the query’s representational requirements. This stage is separated from execution because requirement identification should not be implicitly absorbed into downstream planning. Otherwise, the system may substitute tool-available evidence for query-required evidence, yielding plausible but incomplete formalizations. The Parser avoids this by committing the query to a bounded requirement set before any tool call or computation.

The retrieval follows the dependency order of spatial reasoning: Frame, Entity, State, then Metric. The Parser first resolves the reference frame, then identifies the relevant entities and grounding constraints, next selects the geometric states needed for the queried relation under a coarse-first policy, and finally specifies the metric that directly determines the answer. This makes $\mathcal{P}$ a dependency-consistent requirement set rather than an open-ended factor list. The Parser does not choose tools, instantiate values, or perform computation, which keeps its output independently auditable from execution success. Detailed prompt for Parser is provided in Appendix Table~\ref{tab:prompt_fesm_retrieval}.

\subsection{Executor Agent}
\label{subsec:executor}
The Executor Agent operationalizes the primitives retrieved by the Parser by invoking a curated suite of geometric APIs, as summarized in Table~\ref{tab:api_tools}, bridging the typed FESM requirements with measurable 3D geometry. See Table~\ref{tab:prompt_requirement_guided_tool} for the role prompt and Appendix~\ref{app: toolbox} for module implementation details. 

\textbf{3D Scene Reconstruction.} This module bridges the gap between sparse 2D observations and a unified 3D geometric substrate. Here in our experiments, we adopt VGGT~\citep{wang2025vggt} to estimate camera parameters, depth maps, and generate a unified 3D point cloud. This reconstruction serves as the foundational workspace upon which all subsequent primitive values are grounded. For example, coarse entity positions, orientations, and anchor points are all extracted relative to this shared scene representation, which provides a consistent basis for subsequent deterministic computation.

\textbf{Reference Frame Grounding.} 
The primary challenge of spatial reasoning is not only what relation to compute, but under which reference frame that relation should be interpreted \cite{chen2025geometrically,guo2025pursuing}. We treat this as a reference-frame grounding problem and formalize a reference frame as a 3D coordinate system \(\mathcal{R}\) with origin \(O_\mathcal{R}\) and orthogonal basis vectors \((\mathbf{x}_\mathcal{R}, \mathbf{y}_\mathcal{R}, \mathbf{z}_\mathcal{R})\). In PEPS, the retrieved Frame primitives explicitly determine this coordinate system: \texttt{frame\_type} selects a \textit{camera\_based}, \textit{object\_based}, or \textit{direction\_based} frame, while \texttt{frame\_view}, \texttt{frame\_anchor}, or \texttt{frame\_source}/\texttt{frame\_target} specifies its grounding.
This converts viewpoint and directional ambiguities into checkable and executable commitments.
See Table~\ref{tab:frame_primitives} for details.

\begin{table}[t]
\centering
\caption{Overview of Tool APIs, underlying visual foundation models, and their core skills.}
\label{tab:api_tools}
\resizebox{\textwidth}{!}{%
\begin{tabular}{c|l|p{9.8cm}}
\toprule
\rowcolor{mypink!50}\textbf{API Tools} & \textbf{Visual Models} & \textbf{Skills} \\
\midrule
\texttt{reconstruct} & VGGT~\cite{wang2025vggt} & 3D reconstruction; camera parameters; world points; object alignment \\

\texttt{detect} & GroundingDINO~\cite{liu2024grounding}  & Object detection; bounding boxes; labels \\

\texttt{project\_box\_to\_3d\_points} & SAM-2~\cite{ravi2024sam} & 2D$\rightarrow$3D projection; pixel mask filtering; point extraction \\

\texttt{predict\_obj\_pose} & Orient Anything~\cite{wang2024orient}  & 6-DoF pose estimation; object-to-world transformation; frame setup\\

\texttt{estimate\_scale} & MoGe-2~\cite{wang2025moge} + VGGT & Metric scale computation; align depth maps; real-world units \\

\texttt{ocr} & EasyOCR~\cite{jaided2020easyocr} & Optical character recognition; bounding boxes \\

\texttt{analyze\_motion} & Farneback optical flow~\cite{farneback2003two} & Pixel-level motion; camera movement inference \\

\texttt{code} &  Python sandbox & Python execution \\
\bottomrule
\end{tabular}%
}
\vspace{-1em}
\end{table}

\textbf{Deterministic Program Generation.}
Once the required primitives are instantiated, the Executor invokes the \texttt{code} API for deterministic geometric computation using only $\mathcal{P}$ and the filled workspace, with geometric knowledge from GCA~\cite{chen2025geometrically}.
It cannot request additional perception, introduce undeclared variables, or revise the committed requirements, preventing fallback to open-ended VLM inference as the final stage.
The generation prompt is provided in Table~\ref{tab:prompt_coder_fesm}.

\subsection{Verifier Agent}
\label{subsec:verifier}
The Verifier Agent closes the loop between committed compilation and execution by assessing the candidate $(\hat{a}, \mathcal{T})$ and producing $(v,s,F)$. Rather than solving the query again, it checks whether the answer is fully supported by the query, visual evidence, and primitive-grounded execution trace. If the required primitives are sufficiently instantiated and the computation is supported, $v=\mathrm{accept}$ and execution terminates without further refinement. High-quality accepted traces with $s>\tau$ are stored in the Example Library for future retrieval. Otherwise, the Verifier returns typed feedback that enables targeted repair rather than restarting open-ended reasoning. We distinguish two cases:

\textbf{Missing-slot feedback} indicates that the schema is expressive enough, but some required primitives are missing or insufficiently instantiated, e.g., incomplete frame grounding or metric prerequisites. The Parser then refines only the unresolved requirements using primitives already in the library.

\textbf{Primitive-gap feedback} indicates that the current schema cannot express a query requirement. The Parser may propose temporary primitive hypotheses, which enter the candidate pool and are promoted only after repeated reuse and verified downstream improvement. In this way, the Verifier provides both query-level corrective supervision and schema-level signals for controlled library expansion.
\section{Experiments}
\label{sec:exp}

\begin{table}[t!]
  \caption{\textbf{Comparison with baselines on MMSI-Bench and ViewSpatial-Bench.} The best results are shown in \textbf{bold}, and the second-best are \underline{underlined}. Results for PEPS and MSSR are reported as $\text{Mean}_{\pm \text{Std Dev}}$ over 3 independent runs; other entries are single-run or reported baseline results.} 
  \label{tab:main_results}
  \vspace{-0.6em}
  \centering
    \resizebox{\linewidth}{!}{
    \begin{tabular}{l|ccccc|ccc}

    \toprule

     \multirow{3}{*}{\textbf{Method}}& \multicolumn{5}{c}{\textbf{MMSI-Bench}}
      & \multicolumn{3}{c}{\textbf{ViewSpatial-Bench}} \\
    \cline{2-6} \cline{7-9} 
    & \makecell{Positional\\Relationship} & \makecell{Multi-Step\\Reasoning} & \makecell{Attribute} & Motion & Overall
    & \makecell{Camera\\Based} & \makecell{Person\\Based} & Overall \\

    \midrule
    \midrule
    Random Guessing   & 25.0 & 25.0 & 25.0 & 25.0 & 25.0  &25.5 &27.1 &26.3   \\
    \hline
    % Section for Proprietary LLM models
    \multicolumn{9}{c}{\cellcolor{mypink!50}\emph{\textbf{Proprietary LLMs}}} \\
    o3~\cite{o3}   & 45.8 & 34.9 & 41.6 & 32.0 & 41.0  &51.3 &51.0 &\textbf{51.1}   \\
    
    Gemini 2.5 Pro~\cite{Gemini2.5}    & 39.0 & 34.3 & 36.1 & 33.3 & 36.9 &44.3 &41.6 &43.0 \\
    Gemini 2.5 Flash~\cite{Gemini2.5}   & 37.4& 30.3& 34.5 & 33.4  &35.0&41.6 &36.9 &38.4 \\
    GPT-4o~\cite{openai_gpt4o_2024}   & 28.0 & 30.8 & 32.3 & 36.0 &30.3 & 33.6 & 36.3 & 35.0 \\
    \hline
    \multicolumn{9}{c}{\cellcolor{mypink!50}\emph{\textbf{Open-source VLMs}}} \\
    Llama-3.2-11B-Vision~\cite{Llama-3.2-11B-Vision}        & 26.9 & 19.2 & 20.0 &27.3 &24.5 & 23.7 & 33.6 & 28.8 \\
    LLaVA-OneVision-7B~\cite{LLaVA-OneVision}        & 28.0 & 11.6 & 27.7 & 26.7 &24.5 & 28.5 & 26.6 & 27.5 \\
    Qwen2.5-VL-3B~\cite{Qwen2.5-VL}               & 29.5 & 23.2 & 23.8 & 22.7 &26.5 & 39.8 & 32.1 & 35.8 \\
    Qwen2.5-VL-7B~\cite{Qwen2.5-VL}               & 25.9 & 25.8 & 21.5 & 30.0 &25.9 & 40.6 & 33.4 & 36.9 \\
    Qwen2.5-VL-72B~\cite{Qwen2.5-VL}              & 31.2 & 27.3 & 36.2 & 28.7 &30.7 &47.6 &38.9 &43.1 \\
    {Qwen3-VL-4B}~\cite{yang2025qwen3technicalreport}               &32.8   & 27.8  & 31.0  & 29.7 & 31.1  & 42.3  &35.9  & 39.0 \\
    {Qwen3-VL-8B~\cite{yang2025qwen3technicalreport}   }               & 32.6  & 30.3 & 27.8  & 29.9 & 31.1 & 45.1 & 37.5 & 41.2 \\
    InternVL2.5-8B~\cite{InternVL2.5}              & 29.9 & 30.3 & 27.0 & 24.0 &28.7 & 46.5 & 40.2 & 43.2 \\
    InternVL2.5-2B~\cite{InternVL2.5}               & 30.5 & 26.8 & 29.3 & 26.7 & 29.0 & 32.8 & 37.0 & 35.0 \\
    InternVL3-14B~\cite{InternVL3}               & 25.5 & 29.3 & 26.9 & 28.0 & 39.6 & 47.1 & 33.9 & 40.3 \\
    \hline
\multicolumn{9}{c}{\cellcolor{mypink!50}\emph{\textbf{Trained-based 3D-VLMs}}} \\
{Video-3D-LLM}~\cite{zheng2025video} &25.5  & 25.8 & 22.4 & 25.2 & 25.3 & 24.4 & 35.8 & 30.3 \\
{LLaVA-3D}~\cite{zhu2024llava} &29.5  &19.7  &32.0 & 30.1 &28.0 & 39.6 & 28.3 & 33.8 \\
{VLM-3R}~\cite{fan2025vlm} &33.0  & 30.3 & 31.2 & 31.6 &32.0 & 43.8 & 34.9 & 39.2 \\
VILASR~\cite{wu2025reinforcing}   & 35.9& 23.2& 26.0 & 21.0& 29.8 & 42.4& 34.4&38.2 \\
\hline
    % Section for Agentic models
    \multicolumn{9}{c}{\cellcolor{mypink!50}\emph{\textbf{Tool Integrated Agent}}} \\
    VADAR~\cite{marsili2025visual}          &32.8 &22.7 & 24.3& 27.7&28.9 & 34.2& 33.2&33.7 \\
    pySpatial~\cite{luo2026pyspatial}        &  29.9 & 31.4  &  38.1 & 37.0 &  32.3  & 37.5 & 40.6 &  39.1\\
    GCA~\cite{chen2025geometrically}        & 36.7  & 32.5  & 36.4  & 38.5 &  36.1  & 45.2 & 49.4 & 47.4 \\
    MSSR~\cite{guo2025pursuing}           & \underline{49.4} $\pm0.2$ & \underline{47.8} $\pm0.2$  & \underline{47.1} $\pm0.3$  & \underline{44.9} $\pm0.2$  & \underline{48.1} $\pm0.2$   & \underline{51.0} $\pm0.1$&\textbf{54.4} $\pm0.2$ & \underline{51.8} $\pm0.4$\\

    \rowcolor{blue!5}\textit{\textbf{PEPS (Ours)} }  & \textbf{51.3} $\pm0.2$  & \textbf{48.2} $\pm0.3$   & \textbf{52.6} $\pm0.4$   & \textbf{52.9} $\pm0.4$  & \textbf{51.1} $\pm0.2$    & \textbf{51.9} $\pm0.3$ & \underline{52.9} $\pm0.1$ & \textbf{52.4} $\pm0.2$\\
    
    % MSSR~\cite{guo2025pursuing}           & \underline{49.4}  & \underline{47.8}    & \underline{47.1}    & \underline{44.9}    & \underline{48.1}     & \underline{51.0} &\textbf{54.4} & \underline{51.8} \\

    % \rowcolor{blue!5}\textit{\textbf{PEPS (Ours)} }  & \textbf{51.3}    & \textbf{48.2}     & \textbf{52.6}     & \textbf{52.9}    & \textbf{51.1}      & \textbf{51.9} & \underline{52.9} & \textbf{52.4} \\

    \bottomrule
    \end{tabular}
    }
    \vspace{-1em}
  % \end{adjustbox}

\end{table}

\subsection{Experimental Setup}

\textbf{Benchmarks.} We evaluate PEPS on three complementary benchmarks. \textbf{MMSI-Bench}~\citep{yang2025mmsi} is a manually curated multi-image benchmark for real-world spatial reasoning, covering camera pose, object layout, region relations, and their compositions. Its multi-step subset tests consistent spatial reasoning over longer chains. \textbf{ViewSpatial-Bench}~\citep{li2025viewspatial} focuses on viewpoint-sensitive spatial understanding, particularly egocentric versus allocentric reasoning, across multiple task categories and changing views. We further evaluate on \textbf{OMNI3D-Bench}~\cite{marsili2025visual}, a single-view benchmark for fine-grained 3D spatial reasoning from monocular observations.

\textbf{Baselines.} We compare PEPS against four groups of baselines: (1) \textit{Proprietary LLMs}, including GPT-4o~\cite{openai_gpt4o_2024}, o3~\cite{o3}, and the Gemini-2.5 family~\cite{Gemini2.5}; (2) \textit{Open-source VLMs}, including Llama-3.2-Vision~\cite{Llama-3.2-11B-Vision}, LLaVA-OneVision~\cite{LLaVA-OneVision}, Qwen2.5~\cite{Qwen2.5-VL}/Qwen3-VL~\cite{yang2025qwen3technicalreport}, and InternVL2.5~\cite{InternVL2.5}/InternVL3~\cite{InternVL3}; (3) \textit{Training-based 3D-VLMs}, including Video-3D-LLM~\cite{zheng2025video}, LLaVA-3D~\cite{zhu2024llava}, VLM-3R~\cite{fan2025vlm}, and VILASR~\cite{wu2025reinforcing}, which incorporate explicit 3D representations or 3D-enhanced inference; and (4) \textit{Tool-integrated agents}, including VADAR~\cite{marsili2025visual}, pySpatial~\cite{luo2026pyspatial}, GCA~\cite{chen2025geometrically}, MSSR~\cite{guo2025pursuing}, ViperGPT~\cite{suris2023vipergpt}, and VisProg~\cite{gupta2023visual}.

\textbf{Implementation Details.}
We implement PEPS as a modular multi-agent pipeline, where all three roles are instantiated with GPT-4o through the commercial API, using role-specific prompts and structured JSON interfaces. We set the maximum loop budget to 3 and Example Library threshold $\tau=0.7$. Perceptual tools and geometric computation modules are executed locally. All experiments are conducted on 2 NVIDIA A100 GPUs. We reproduce baseline results using their officially released code and evaluation settings. Full hyperparameter choices are provided in Appendix~\ref{app:implementation}.

\subsection{Main Result}
\myfinding{PEPS generalizes strongly across benchmark settings} On MMSI-Bench, PEPS reaches \textbf{51.1\%} overall accuracy, exceeding the strongest proprietary baseline o3 at 41.0\% by 10.1 points and outperforming recent tool-integrated methods such as GCA at 36.1\% and MSSR at 48.1\%. This advantage remains stable on ViewSpatial-Bench, where PEPS achieves a new best overall score of \textbf{52.4\%}, surpassing GCA at 47.4\% and MSSR at 51.8\%. The same pattern carries over to OMNI3D-Bench, where PEPS obtains the best overall result of \textbf{49.0\%}. Taken together, these results show that the gain of PEPS is not tied to a single benchmark format, but transfers across multi-image, viewpoint-sensitive, and monocular 3D spatial reasoning settings.

\begin{wraptable}{l}{0.62\linewidth}
\vspace{-1.4em}
\caption{\textbf{Performance comparison on OMNI3D-Bench}.}
\label{tab:omni3d}
\vspace{-0.5em}
\centering
\footnotesize
\setlength{\tabcolsep}{3pt}

\resizebox{\linewidth}{!}{%
\begin{tabular}{l|ccccc}
\toprule
\textbf{Method} & Counting & Float-MRA & Yes/No & Multiple Choice & \textbf{Overall} \\
\midrule
\rowcolor{mypink!50}
\multicolumn{6}{l}{\textit{Generic VLMs}} \\
GPT-4o & 28.1 & 35.5 & 66.7 & 57.2 & 42.9 \\
Claude3.5-Sonnet & 22.4 & 20.6 & 62.2 & 50.6 & 32.2 \\
Llama-3.2 & 24.3 & 19.3 & 47.5 & 27.4 & 25.6 \\
Gemini1.5-Pro & 25.2 & 28.1 & 46.2 & 37.6 & 32.0 \\
Gemini1.5-Flash & 24.3 & 27.6 & 51.1 & 52.9 & 35.0 \\
Qwen2-VL-7B-Instruct & 12.9 & 21.5 & 58.7 & 33.7 & 28.0 \\
LLaVA-OV-7B-Chat & 22.9 & 26.8 & 60.0 & 27.9 & 31.4 \\
SpaceMantis & 20.0 & 21.7 & 50.6 & 48.2 & 30.3 \\
\midrule
\rowcolor{mypink!50}
\multicolumn{6}{l}{\textit{Visual Programming Approaches}} \\
ViperGPT & 20.0 & 15.4 & 56.0 & 42.4 & 26.7 \\
VisProg & 2.9 & 0.9 & 54.7 & 25.9 & 13.5 \\
VADAR & 21.7 & 35.5 & 56.0 & 57.6 & 40.4 \\
pySpatial & 22.9 & 38.6 & 72.0 & 54.7 & 44.2 \\
GCA & 25.6 & 40.2 & 70.1 & \underline{59.3} & 45.9 \\
MSSR & \underline{28.4} $\pm0.1$ & \underline{42.7} $\pm0.3$ & \textbf{73.8} $\pm0.5$ & 59.1 $\pm0.4$ & \underline{48.2} $\pm0.4$\\
\rowcolor{blue!5}
\textit{\textbf{PEPS (Ours)}} &
\textbf{28.8} $\pm0.4$ & \textbf{44.0} $\pm0.7$ & \underline{73.5} $\pm0.2$ &
\textbf{59.5} $\pm0.4$ & \textbf{49.0} $\pm0.5$ \\
\bottomrule
\end{tabular}%
}

\vspace{-0.8em}
\end{wraptable}

\myfinding{PEPS is especially effective on viewpoint-sensitive and computation-heavy reasoning} On MMSI-Bench, PEPS performs best on Positional Relationship, Multi-Step Reasoning, and Motion, reaching 51.3\%, 52.6\%, and 52.9\%, respectively, suggesting that executable primitive parsing benefits both basic localization and compositional reasoning. On ViewSpatial-Bench, it performs consistently across Camera Based and Person Based settings (51.9\% and 52.9\%), maintaining stable spatial commitment across externally anchored and human-centered frames. On OMNI3D-Bench, PEPS achieves 44.0\% on Float-MRA, outperforming GCA (40.2\%) and MSSR (42.7\%) on precise metric computation.

\begin{table}[t]
\scriptsize
\renewcommand{\arraystretch}{1.2}
\setlength{\tabcolsep}{3pt}
\centering

\caption{
\textbf{Average Per-query API Usage and Cost on MMSI-Bench.}
}
\label{tab:api_usage_cost}
\vspace{-1em}
\resizebox{\linewidth}{!}{%
\begin{tabular}{l c c c c c c}
\toprule
\textbf{Method}
& \makecell{\textbf{API Cost}\\\textbf{(USD/query)}}
& \makecell{\textbf{API Calls}\\\textbf{(calls/query)}}
& \makecell{\textbf{Non-cached Input}\\\textbf{(tokens/query)}}
& \makecell{\textbf{Cached Input}\\\textbf{(tokens/query)}}
& \makecell{\textbf{Total Input}\\\textbf{(tokens/query)}}
& \makecell{\textbf{Output}\\\textbf{(tokens/query)}} \\
\midrule

MSSR (GPT-4o)
& 0.1800
& 4.30
& 7,200.00
& 1,600.00
& 8,800.00
& 16,000.00 \\

PEPS (GPT-4o)
& 0.1530 \down{15.0\%}
& 3.60 \down{16.3\%}
& 6,000.00 \down{16.7\%}
& 4,800.00 \up{200.0\%}
& 10,800.00 \up{22.7\%}
& 13,200.00  \down{17.5\%}\\

\midrule

GPT-5
& 0.0862
& 1.00
& 4,900.00
& 900.00
& 5,800.00
& 8,000.00 \\

MSSR (GPT-5)
& 0.1562
& 3.90
& 6,400.00
& 1,800.00
& 8,200.00
& 14,800.00 \\

PEPS (GPT-5)
& 0.1231 \down{21.2\%}
& 3.30 \down{15.4\%}
& 5,200.00 \down{18.8\%}
& 4,800.00 \up{166.7\%}
& 10,000.00 \up{22.0\%}
& 11,600.00  \down{21.6\%}\\

\midrule

Gemini 3.6 Flash
& 0.0557
& 1.00
& 4,500.00
& 1,000.00
& 5,500.00
& 6,500.00 \\

MSSR (Gemini 3.6 Flash)
& 0.1005
& 3.64
& 5,800.00
& 2,200.00
& 8,000.00
& 12,200.00 \\

PEPS (Gemini 3.6 Flash)
& 0.0769 \down{23.5\%}
& 3.24 \down{11.0\%}
& 4,800.00 \down{17.2\%}
& 4,700.00 \up{113.6\%}
& 9,500.00 \up{18.8\%}
& 9,200.00 \down{24.6\%}\\

\bottomrule
\end{tabular}
}
\end{table}

\begin{table}[t]
\scriptsize
\renewcommand{\arraystretch}{1.2}
\setlength{\tabcolsep}{3pt}
\centering

\caption{
\textbf{Average Per-query Computational Overhead on MMSI-Bench.}
}
\label{tab:computational_overhead}
\vspace{-1em}
\resizebox{\linewidth}{!}{%
\begin{tabular}{l c c c c c c c c}
\toprule
\textbf{Method}
& \makecell{\textbf{GPU Memory}\\\textbf{(GB/process)}}
& \makecell{\textbf{Mean E2E}\\\textbf{Latency}\\\textbf{(s/query)}}
& \makecell{\textbf{Median E2E}\\\textbf{Latency}\\\textbf{(s/query)}}
& \makecell{\textbf{P95 E2E}\\\textbf{Latency}\\textbf{(s/query)}}
& \makecell{\textbf{Average}\\\textbf{Iterations}\\\textbf{(rounds/query)}}
& \makecell{\textbf{1-Iteration}\\\textbf{(\%)}}
& \makecell{\textbf{2-Iteration}\\\textbf{(\%)}}
& \makecell{\textbf{3-Iteration}\\\textbf{(\%)}} \\
\midrule

MSSR (GPT-4o)
& 7.20
& 86.75
& 79.60
& 137.40
& 2.15
& 15.0\%
& 55.0\%
& 30.0\% \\

PEPS (GPT-4o)
& 9.80 \up{36.1\%}
& 53.60 \down{38.2\%}
& 46.20 \down{42.0\%}
& 92.80 \down{32.5\%}
& 1.20 \down{44.2\%}
& 83.0\% \up{68.0\%}
& 14.0\% \down{41.0\%}
& 3.0\% \down{27.0\%} \\

\midrule

GPT-5
& 0.00$\dagger$
& 34.80
& 31.20
& 60.50
& 1.00
& 100.0\%
& 0.0\%
& 0.0\% \\

MSSR (GPT-5)
& 7.20
& 98.75
& 96.40
& 165.20
& 1.95
& 25.0\%
& 55.0\%
& 20.0\% \\

PEPS (GPT-5)
& 9.80 \up{36.1\%}
& 66.83 \down{32.3\%}
& 61.50 \down{36.2\%}
& 117.80 \down{28.7\%}
& 1.10 \down{43.6\%}
& 91.5\% \up{66.5\%}
& 7.0\% \down{48.0\%}
& 1.5\% \down{18.5\%}\\

\midrule

Gemini 3.6 Flash
& 0.00
& 23.40
& 20.80
& 42.60
& 1.00
& 100.0\%
& 0.0\%
& 0.0\% \\

MSSR (Gemini 3.6 Flash)
& 7.20
& 72.40
& 67.90
& 125.60
& 1.82
& 30.0\%
& 58.0\%
& 12.0\% \\

PEPS (Gemini 3.6 Flash)
& 9.80 \up{36.1\%}
& 45.70 \down{36.9\%}
& 40.80 \down{39.9\%}
& 82.40 \down{34.4\%}
& 1.08 \down{40.7\%}
& 93.0\% \up{63.0\%}
& 6.0\% \down{52.0\%}
& 1.0\% \down{11.0\%}\\

\bottomrule
\end{tabular}
}
\vspace{-2em}
\end{table}

% \vspace{-0.4em}
\subsection{Parse Before Execution Reduces Reasoning Overhead}
\myfinding{PEPS achieves higher accuracy with substantially less iterative reasoning overhead}
As shown in Tables~\ref{tab:api_usage_cost} and~\ref{tab:computational_overhead}, PEPS reduces API cost by 15.0-23.5\% and mean end-to-end latency by 32.3-38.2\% over MSSR across GPT-4o, GPT-5, and Gemini 3.6 Flash.
This efficiency gain follows from the \textit{parse-before-execute} design: by committing to executable requirements before tool invocation, 83.0-93.0\% of queries terminate after one round, reducing average refinement from 1.82-2.15 rounds for MSSR to 1.08-1.20 for PEPS.
Although PEPS uses more cached context and GPU memory, fewer API calls and output tokens yield lower practical inference cost and latency.
Together with the consistent accuracy gains, these results show that explicit requirement commitment enables a stronger accuracy-efficiency trade-off rather than relying on repeated downstream refinement.

\subsection{Ablation Study}
\label{sec:ablation}

\begin{figure}[t!] % [htbp]
    \centering
    \includegraphics[width=1.0\textwidth]{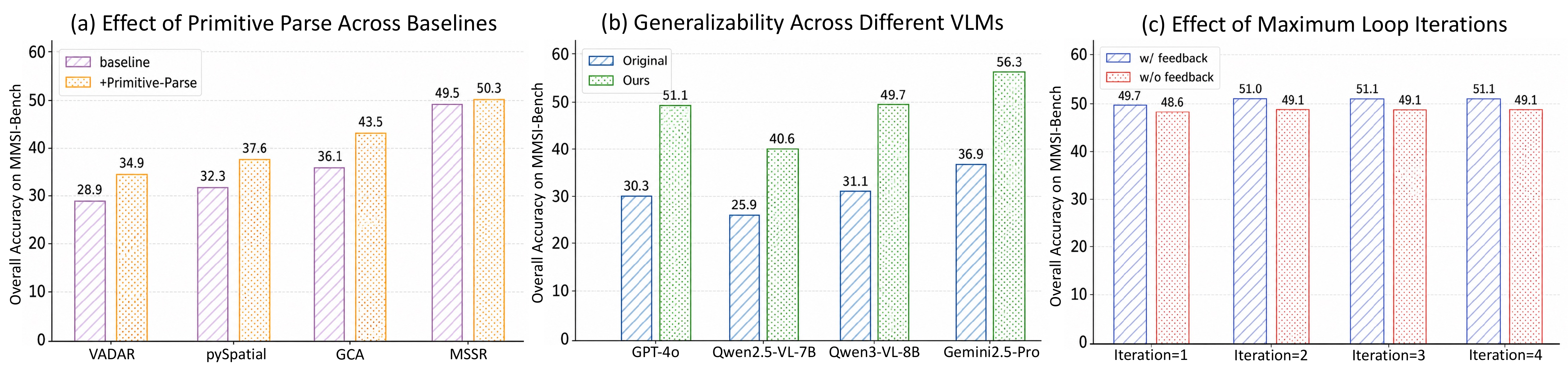} 
    \vspace{-1.8em}
    \caption{\textbf{Ablation studies.} We evaluate three scenarios: (a) prepending primitive parsing to existing spatial reasoning agentic baselines; and (b) applying PEPS to different VLM backbones; and (c) varying the maximum loop budget, with or without corrective Verifier feedback.} 
    \label{fig:ablation}
    \vspace{-2em}
\end{figure}

\textbf{Effect of Spatial Primitives.}
We examine whether the gains come from the FESM primitive interface itself. Table~\ref{tab:primitive_ablation} compares PEPS with three weakened variants: Free-form uses natural-language descriptions; Flat-schema removes the Frame, Entity, State, and Metric hierarchy; and Fixed-schema disables Verifier-driven schema updates. All three variants perform worse than the full model. The largest drop comes from free-form parsing, showing that open-ended requirement descriptions are less reliable than executable primitive commitments. The flat-schema drop shows that the FESM hierarchy provides useful dependency structure, while the fixed-schema drop suggests that controlled schema expansion improves coverage. We also prepend the same primitive parsing stage to baselines while keeping their downstream pipelines unchanged; all baselines improve as shown in Figure~\ref{fig:ablation}(a). This indicates that FESM parsing acts as a general committed-compilation front end.

\begin{table}[t]
    \centering
    \begin{minipage}{0.34\textwidth}
        \centering
        \setlength{\abovecaptionskip}{2pt}
        \caption{Ablation study on spatial primitives. “Flat-schema” removes the FESM hierarchy, and “Fixed-schema” disables the primitive-pool update mechanism.}
        \resizebox{\linewidth}{!}{
        \begin{tabular}{@{}lcc@{}}
            \toprule
{Method} & MMSI  & OMNI3D \\
            \midrule
            GPT-4o          & 30.3 & 42.9 \\
            \rowcolor{gray!10}Ours (Full)      & \textbf{51.1} & \textbf{49.0} \\
            \midrule
            Free-form         & 42.4 & 44.8 \\
            Flat-schema         & 49.2 & 47.7 \\
            Fixed-schema          & 50.7 & 48.5 \\
            \bottomrule
        \end{tabular}
        }
        \label{tab:primitive_ablation}
    \end{minipage}
    \hspace{0.02\textwidth} 
    \begin{minipage}{0.62\textwidth}
    \setlength{\abovecaptionskip}{2pt}
    \caption{Ablation study on the contribution of each component. “Multi-Agent” denotes explicit role decomposition into three sub-tasks, “Feedback” denotes corrective feedback returned by the Verifier, and “Example” denotes retrieval from the Example Library of high-quality historical candidates.}
        \centering
        \resizebox{\linewidth}{!}{
       \begin{tabular}{cccc|c}
    \toprule
   Multi-Agent & Verifier & Feedback & Example  & MMSI-Bench \\
    \midrule
    $\usym{2717}$ & \checkmark  & \checkmark  & \checkmark  & 49.2 \\
    \checkmark    & $\usym{2717}$ & $\usym{2717}$ & $\usym{2717}$ & 47.6 \\
    \checkmark    & \checkmark    & $\usym{2717}$ & $\usym{2717}$ &  48.9 \\
    \checkmark    & \checkmark    & \checkmark    & $\usym{2717}$ & 50.4 \\
    \midrule
    \rowcolor{gray!10} \checkmark    & \checkmark    & \checkmark    & \checkmark    & \textbf{51.1} \\

    \bottomrule
\end{tabular}
        }
        \label{tab:component_ablation}
    \end{minipage}
    % \vspace{-1em}
\end{table}
\textbf{Component Contribution.} Table~\ref{tab:component_ablation} illustrates the contribution of each core component by progressively weakening the full system along four axes: multi-agent role decomposition, verification, corrective feedback, and example-based experience retrieval. The largest weakness appears when verification is removed, indicating that PEPS relies on trace-level inspection to prevent incomplete primitive commitments from passing into the final answer. Corrective feedback is also critical because it turns verification from passive filtering into active repair, allowing the Parser to recover missing slots instead of merely retrying. Role separation improves reliability by keeping requirement parsing, value acquisition, and answer checking from collapsing into one loosely constrained reasoning step.

\begin{wraptable}{r}{0.36\linewidth}
\vspace{-1em}
\captionsetup{skip=2pt}
\centering
\caption{Cross-Dataset Generalizability of Primitive Schema.}
\resizebox{\linewidth}{!}{%
\begin{tabular}{lcc}
\toprule
Method & MindCube-Tiny & OmniSpatial \\
\midrule
GPT-4o & 35.8 & 51.5 \\
Free-form & 61.3 & 62.1\\
PEPS & 73.7 & 69.8\\
\bottomrule
\end{tabular}
}
\label{tab:cross dateset}
\vspace{-1em}
\end{wraptable}

\textbf{Cross-Dataset Generalizability of Primitive Schema.}
We test whether the primitive schema transfers to unseen benchmarks.
As shown in Table~\ref{tab:cross dateset}, the fixed source-induced schema consistently outperforms free-form parsing on MindCube-Tiny~\cite{yin2025spatial} and OmniSpatial~\cite{jia2025omnispatial}.
This suggests that FESM captures reusable execution requirements rather than benchmark-specific cues, remaining effective as scene composition and query style shift across datasets.
Importantly, this transfer requires no benchmark-specific schema reconstruction, supporting FESM as a reusable interface.

\begin{wraptable}{r}{0.36\linewidth}
\vspace{-1em}
\captionsetup{skip=2pt}
\centering
\small
\caption{Sensitivity analysis of different 3D reconstruction backbones.}
% \resizebox{\linewidth}{!}{%
\begin{tabular}{lc}
\toprule
Method & MMSI-Bench  \\
\midrule
GPT-4o           & 30.3  \\
PEPS w/ VGGT   & 51.1  \\
PEPS w/ $\pi^3$    & 51.8  \\
PEPS w/ CUT3R  & 50.5  \\
\bottomrule
\end{tabular}
% }
\label{tab:ablation_3Dbackbone}
\vspace{-1em}
\end{wraptable}

\textbf{Sensitivity Analysis on 3D Reconstruction Backbones.} We evaluate whether PEPS depends on a specific 3D reconstruction backbone by replacing VGGT with $\pi^3$~\cite{wang2025pi} and CUT3R~\cite{wang2025continuous}, with others unchanged. As shown in Table~\ref{tab:ablation_3Dbackbone}, all reconstruction backbones substantially outperform the GPT-4o baseline, indicating that the main gain does not depend on a particular reconstruction model. $\pi^3$ gives the best result, while VGGT and CUT3R remain close. This suggests that PEPS is robust to the geometric substrate as long as the reconstruction provides usable camera poses and object-level 3D states.

\textbf{Generalizability Across VLM Backbones.} Figure~\ref{fig:ablation}(b) suggests that PEPS functions as model-agnostic rather than a prompt tuned to one particular backbone. Across different VLMs, the primitive interface consistently provides a more stable reasoning scaffold by narrowing each model’s role from open-ended spatial inference to structured requirement retrieval, grounding, and verification. This is especially important for weaker backbones, where unconstrained planning is more likely to drift, but it also benefits stronger models by reducing ambiguity in reference frames and metric selection.

\textbf{Effect of Iteration Budget.}
We study the effect of the maximum loop budget by varying the number of refinement rounds with and without corrective feedback. As displayed in Figure~\ref{fig:ablation}(c), performance improves mainly in the early rounds and quickly saturates. The gap between the feedback and no-feedback setting shows that the loop is not simply benefiting from repeated attempts, but from targeted repair of incomplete primitive commitments. And the limited gain from later rounds suggests that remaining failures are less likely to be fixed by further parsing refinement and more likely to stem from perception errors, reconstruction noise, or genuine schema gaps. This indicates that PEPS functions as a fast diagnostic repair loop rather than an open-ended self-refinement process.
\begin{figure}[t!] % [htbp]
    \centering
    \includegraphics[width=1.0\textwidth]{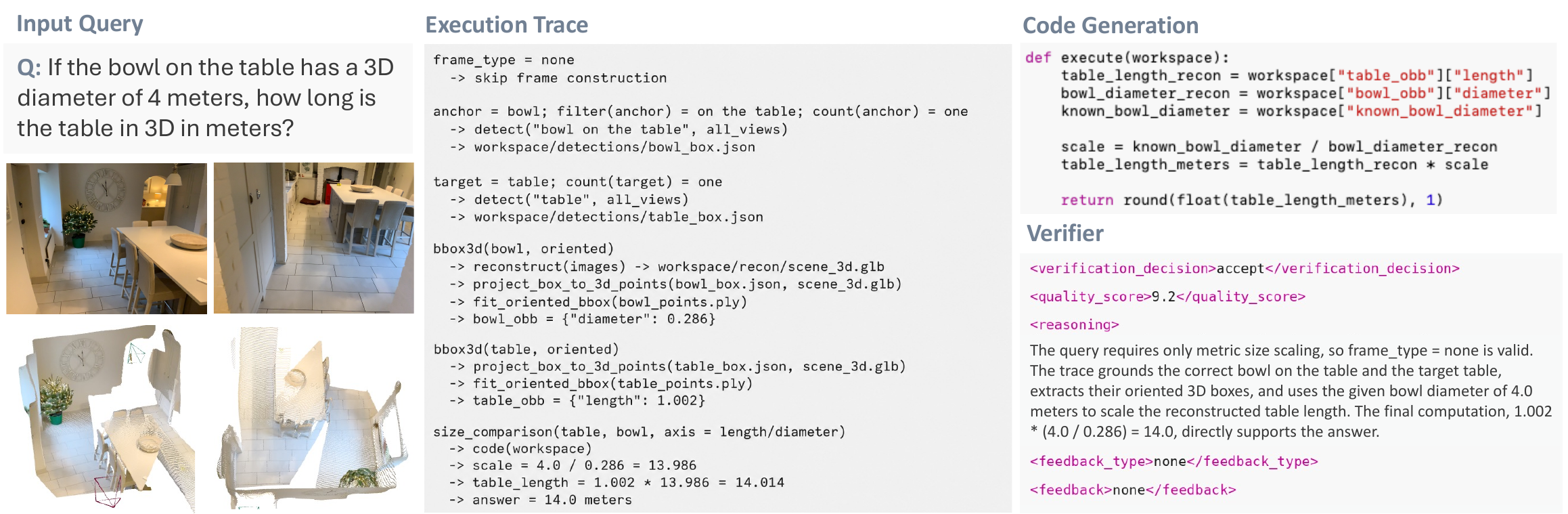} 
    % \vspace{-1.8em}
    \caption{\textbf{Compact Qualitative Example.} More detailed examples are provided in Appendix~\ref{app:qualitativeexample}.} 
    \label{fig:example}
    \vspace{-1em}
\end{figure}
\vspace{-0.6em}
\subsection{Failure Cases Analysis}
\begin{wrapfigure}{r}{0.26\linewidth}
\vspace{-1em}
\captionsetup{skip=2pt}
\centering
\includegraphics[width=\linewidth]{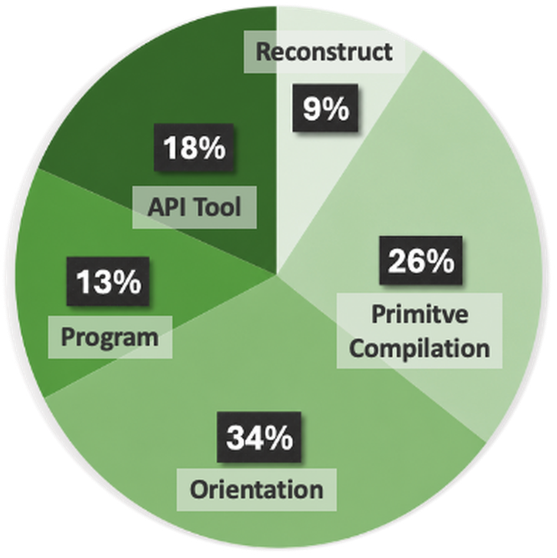}
\caption{Failure cases distribution.}
\label{fig:failure}
\vspace{-2em}
\end{wrapfigure}
We manually analyze about 100 failed examples and attribute each to its earliest dominant error source, as shown in Figure~\ref{fig:failure}. Notably, most failures arise before final deterministic computation. Primitive Compilation errors (26\%) result from incomplete or mis-specified primitive sets that lead execution toward the wrong formalization. Orientation errors (34\%), the largest category, show that grounding object-centric or event-conditioned directions remains the main bottleneck. API Tool errors (18\%) arise from incorrect parameter passing, tool sequencing, unresolved dependencies, or budget exhaustion. Program errors (13\%) are less frequent and are mostly caused by mistakes in deterministic geometric computation, such as missing coordinate transformations or incorrect metric mapping. Reconstruction errors (9\%) reflect failures in the 3D scene geometry, where inaccurate poses or point clouds prevent reliable primitive instantiation. 

\vspace{-1.2em}
\section{Related Work}
\label{sec:relatedwork}
\vspace{-0.2em}

Recent spatial reasoning methods move beyond direct VLM inference through external tools~\cite{wu2025transductive, marsili2025visual}, visual intermediate reasoning~\cite{wu2025reinforcing}, and explicit 3D representations~\cite{fan2025vlm, zheng2025video}. Visual programming and agentic methods further decompose questions into executable code, tool plans, dynamic APIs, or geometric computation~\cite{gupta2023visual, suris2023vipergpt, luo2026pyspatial}. While these approaches improve grounding and compositionality, their query-level artifacts often remain open-ended, mixing perception, planning, and answer derivation after the query is transformed into a program, plan, or formalization. A complementary line studies minimal sufficient evidence, showing that task-relevant observations can improve reliability and reduce distractors~\cite{guo2025pursuing}. Together, these works suggest that spatial reasoning requires not only stronger tools or richer geometry, but tighter control over what information is committed before computation. See Appendix~\ref{app:relatedwork} for full related work.
\vspace{-0.8em}
\section{Conclusion}
\label{sec:conclusion}
\vspace{-0.2em}
We present PEPS, a training-free agentic framework that addresses under-committed query compilation in spatial reasoning by parsing queries into executable primitives under a typed FESM schema. By separating primitive parsing, geometric execution, and trace-level verification, PEPS makes reasoning commitments explicit, bounded, and diagnosable. Experiments across multi-image MMSI-Bench, viewpoint-sensitive ViewSpatial-Bench, and monocular OMNI3D-Bench show consistent improvements over strong VLM and tool-integrated baselines. More broadly, our results suggest that reliable spatial reasoning depends not only on stronger perception tools or richer 3D geometry, but also on explicitly committing to the right representational requirements before computation begins.

\section*{AI Use Statement}

In this work, we used generative AI tools solely for language editing and writing polish, including grammar checking, wording refinement, and readability improvements. We did not use generative AI tools to develop the research methodology, design experiments, interpret results, or formulate scientific claims. All AI-assisted edits were reviewed by the authors. We take full responsibility for the final content of this work.

\bibliography{iclr2027_conference}
\bibliographystyle{iclr2027_conference}

\appendix
\newpage

\section{Implementation Details}
\label{app: toolbox}

\subsection{Visual Foundation Models}
The Executor instantiates the parsed primitives through a set of off-the-shelf visual foundation models. 
These models are used as perception backends for filling primitive values, such as camera pose, object location, object extent, metric scale, and orientation. 
The subsequent spatial relation is computed deterministically from the instantiated variables rather than inferred directly by these models.

\begin{itemize}
    \vspace{+0.5mm}
    \item \textbf{VGGT}~\cite{wang2025vggt}.
    We use VGGT to build the shared 3D workspace from the input image set.
    Given one or multiple views, it estimates camera parameters, depth maps, and point maps, which provide the common geometric substrate for grounding camera-based frames, object centroids, and cross-view spatial relations.
    In our pipeline, VGGT supplies the reconstruction context on which later primitive values are instantiated.

    \vspace{+0.5mm}
    \item \textbf{MoGe-2}~\cite{wang2025moge}.
    MoGe-2 is used when a query requires metric-scale reasoning.
    It predicts monocular geometry with scale information, allowing the relative 3D reconstruction to be aligned with real-world units.
    We use this scale estimate to support primitives and metrics involving physical size, distance, or measurements in meters.

    \vspace{+0.5mm}
    \item \textbf{GroundingDINO}~\cite{liu2024grounding}.
    We use GroundingDINO for open-vocabulary entity localization.
    Given textual entity descriptions from the parsed Entity primitives, it returns candidate boxes for objects, regions, or referring expressions in the specified view.
    These detections serve as the entry point for grounding anchors, targets, and candidate sets before acquiring their 3D states.

    \vspace{+0.5mm}
    \item \textbf{SAM-2}~\cite{ravi2024sam}.
    SAM-2 refines detected boxes into object-level masks.
    We use these masks to select the corresponding 3D points from the reconstructed workspace, thereby converting 2D detections into entity-level point sets.
    This bridge is needed for state primitives such as point clouds, centroids, and 3D extents.

    \vspace{+0.5mm}
    \item \textbf{Orient Anything}~\cite{wang2024orient}.
    Orient Anything provides object orientation estimates from a single view.
    We use its predicted pose or principal direction to instantiate orientation-related State primitives and to construct object-based reference frames when a query depends on an object's facing direction, surface normal, or local coordinate system.
\end{itemize}

\subsection{Tool Interfaces}
\label{subsec:app_tool_interfaces}
\begin{itemize}
    \vspace{+0.5mm}
    \item \texttt{reconstruct}. 
    This API builds the shared 3D workspace from the input image set. 
    Given one or multiple views, it invokes VGGT to estimate camera intrinsics, camera extrinsics, depth, and world-space point maps. 
    For multi-view inputs with possible scene variation, the pipeline first identifies stable overlapping regions or common static objects for alignment before constructing the unified reconstruction. 
    The resulting workspace provides the geometric substrate for camera-based frames, cross-view grounding, and object-level state acquisition.

    \vspace{+0.5mm}
    \item \texttt{detect}. 
    This API localizes entities specified by the parsed Entity primitives. 
    It takes a text description and a target view as input, and returns candidate bounding boxes with labels. 
    Depending on the deployed backbone, localization is performed either through a capable VLM prompted for region grounding or through GroundingDINO for open-vocabulary detection.

    \vspace{+0.5mm}
    \item \texttt{project\_box\_to\_3d\_points}. 
    This API converts a localized 2D entity into a 3D entity state. 
    Starting from a detected bounding box, it obtains an object mask with SAM-2 and uses the mask to select corresponding points from the reconstructed world-space point map. 
    The output is an entity-level point set, from which centroids, extents, and surface-level quantities can be derived.

    \vspace{+0.5mm}
    \item \texttt{predict\_obj\_pose}.
    This API estimates the local pose of an object when a query requires orientation-sensitive reasoning. 
    It combines the object's 3D location from the projected point set with an orientation estimate from Orient Anything, and returns an object-to-world transform. 
    The transform is used to instantiate orientation states and object-based reference frames.

    \vspace{+0.5mm}
    \item \texttt{estimate\_scale}.
    This API is invoked for queries involving metric distance, physical size, or real-world units. 
    It aligns the relative reconstruction produced by VGGT with the metric geometry estimated by MoGe-2, producing a global scale factor that maps reconstructed coordinates to meters.

    \vspace{+0.5mm}
    \item \texttt{ocr}.
    This API recognizes text in an image and returns detected strings with their bounding regions. 
    We use it when textual cues are part of entity grounding, region filtering, or answer candidate identification.

    \vspace{+0.5mm}
    \item \texttt{analyze\_motion}.
    This API estimates image-level motion between sequential views using Farneback optical flow~\cite{farneback2003two}. 
    It is used for motion-sensitive queries where small viewpoint changes or local camera displacement may not be reliably captured by full 3D reconstruction alone.

    \vspace{+0.5mm}
    \item \texttt{code}.
    This API performs the final deterministic computation over the instantiated primitive values. 
    It receives the filled workspace variables, the retrieved FESM requirements, and the relevant geometric rules, then generates and executes sandboxed Python code. 
    The code is constrained to use only available variables, preventing the computation stage from inventing new perceptual evidence or revising the committed requirements.
\end{itemize}

\subsection{Agentic Framework}
\label{subsec:app_framework}
We implement the system as a modular pipeline that separates VLM-based reasoning from perception and geometric computation. This separation allows the Parser, Executor, and Verifier to share a common state while invoking visual tools only when required by the current FESM commitments.

\noindent\textbf{State and Orchestration.}
The pipeline is implemented with Ray~\cite{moritz2018ray} and LangGraph. LangGraph maintains the intermediate workspace, including parsed primitives, tool outputs, execution traces, and verifier feedback, and coordinates the iterative loop.

\noindent\textbf{Tool Deployment.}
Perception and computation modules are served as independent Ray Serve actors. This design supports parallel tool calls, such as reconstruction and detection, and isolates failures from individual visual backends. In our experiments, the tool suite runs on 2 NVIDIA A100 GPUs.

\noindent\textbf{VLM Deployment.}
VLM roles are deployed separately from the tool services. Open-source backbones are served with vLLM~\cite{kwon2023efficient} on 4 NVIDIA A100 GPUs, while closed-source models are accessed through their official APIs. We use role-specific decoding: the Parser, Executor, and Verifier use \texttt{TEMPERATURE=0.6} and \texttt{TOP\_P=0.95} to allow flexible requirement analysis and feedback, whereas the code generation stage uses \texttt{TEMPERATURE=0.0} for deterministic computation.

\subsection{Implementation}
\label{app:implementation}

\noindent\textbf{Agent Configuration.}
Unless otherwise specified, the Parser, Executor, and Verifier are instantiated with GPT-4o. 
For open-source backbone experiments, we replace the VLM used by all agent roles with the evaluated backbone while keeping the tool suite, prompts, and execution protocol unchanged. 
We use \texttt{TEMPERATURE=0.6} and \texttt{TOP\_P=0.95} for requirement parsing, tool orchestration, and verification, and set \texttt{TEMPERATURE=0.0} for code generation to reduce stochastic variation in geometric computation. 
Each query is allowed at most $3$ parsing--execution--verification rounds.

\noindent\textbf{Schema and Example Library.}
The initial FESM schema is induced from $300$ sampled training/development questions, with $100$ questions from each source benchmark. 
Candidate primitives are merged by semantic equivalence and retained only when they are supported by at least $3$ questions and can be grounded by the available tool interface. 
During execution, verifier feedback is categorized as either \emph{missing-slot} or \emph{primitive-gap}. 
Missing-slot feedback triggers re-parsing within the existing schema, while primitive-gap feedback adds a temporary primitive candidate to the pool. 
A candidate primitive is promoted to the schema after being reused in at least $5$ accepted cases with verifier score above $\tau=0.7$. 
The Example Library stores accepted traces, including the query, parsed primitives, instantiated variables, generated program, final answer, and verifier feedback. 
For each new query, we retrieve the top-$3$ examples by embedding similarity and use them as in-context demonstrations. 
Ground-truth answers from test sets are never used for schema induction, example retention, or verifier feedback.

\noindent\textbf{Execution Protocol.}
The Executor issues tool calls only to instantiate unresolved FESM requirements. 
Each query has a maximum budget of $12$ perception/tool calls and $3$ code-generation attempts. 
Reconstruction is cached per input scene, and entity detections are cached by view and text prompt. 
If a tool returns multiple candidates, the Executor applies the parsed view, region, attribute, and cardinality constraints for disambiguation; if no valid candidate remains, the case is marked as unresolved and returned to the Verifier. 
Parallelizable calls, such as detecting multiple entities after reconstruction, are executed concurrently.

\noindent\textbf{Code Execution and Answer Normalization.}
Generated Python code runs in a sandbox with access only to the filled workspace variables and standard numerical libraries such as \texttt{numpy}, \texttt{scipy}, and \texttt{math}. 
The code stage cannot request new perception results, introduce undeclared variables, or modify the parsed requirements. 
Each execution is limited to $10$ seconds. 
For multiple-choice questions, the output is mapped to the closest valid option; for yes/no questions, it is normalized to a binary label; for directional answers, synonyms are canonicalized to the benchmark label set; and for numerical answers, units and rounding are normalized according to the official evaluation protocol.

\noindent\textbf{Evaluation Setup.}
All benchmarks are converted to a unified input format consisting of images, view identifiers, question text, and optional answer candidates. 
Multi-image inputs preserve their original view order, while single-image 3D tasks are treated as one-view scenes. 
We use the original benchmark questions without ground-truth intermediate supervision. 
Baselines are evaluated using their official code and recommended settings whenever available; otherwise, we follow the closest released prompt and inference configuration. 
All experiments are run on NVIDIA A100 GPUs with the perception tool suite deployed separately from the VLM inference service.
\section{Related Work}
\label{app:relatedwork}
\noindent\textbf{Spatial Reasoning in Vision-Language Models.}
Spatial reasoning requires models to interpret object relations, reference frames, viewpoint changes, and 3D scene structure, and remains a persistent challenge for modern VLMs~\cite{yang2025thinking, yu2025far, yang2025mmsi, li2025viewspatial, yin2025spatial, jia2025omnispatial}. Recent work improves spatial ability through stronger multimodal backbones, spatial benchmarks, specialized instruction tuning, and geometry-aware representations~\cite{hurst2024gpt,openai_gpt4o_2024, LLaVA-OneVision, Qwen2.5-VL, qwenteam2025qwen3vl, cheng2024spatialrgpt, ma2025spatialllm}. These approaches narrow the gap between 2D visual recognition and 3D spatial understanding by injecting depth, reconstruction, position-aware features, or embodied observations. However, learned spatial representations alone do not necessarily expose whether the model has selected the correct viewpoint, grounded the intended entities, or acquired the geometric variables needed by the query.

\vspace{+0.5mm}
\noindent\textbf{Tool-Integrated and Geometric Spatial Reasoning.}
Another line of work moves spatial reasoning beyond direct VLM inference by introducing executable intermediate representations, external perception tools, dynamic APIs, and deterministic geometric computation~\cite{gupta2023visual, suris2023vipergpt, wu2025transductive, marsili2025visual, luo2026pyspatial, chen2025geometrically, fan2025vlm}. Program- and agent-based systems improve compositionality by translating visual questions into code, tool plans, or modular reasoning traces. 3D-geometric systems further ground inference with reconstructed scenes, camera poses, object-level geometry, and metric computation. While these methods reduce reliance on unconstrained semantic reasoning, their generated artifacts often remain open-ended, mixing perception, planning, and answer derivation in one output space. Thus, plausible execution traces may still hide missing spatial commitments, such as the intended reference frame, entity constraint, state variable, or relation metric.

\vspace{+0.5mm}
\noindent\textbf{Evidence Selection and Constrained Reasoning.}
A complementary direction studies task-relevant evidence and minimal sufficiency, showing that spatial reasoning can benefit from selecting only the observations necessary for the query~\cite{guo2025pursuing}. More broadly, constrained reasoning restricts model behavior by translating natural language into explicit variables, constraints, formal programs, or verifiable symbolic structures before solving~\cite{hitzler2022neuro, pan2023logic, ye2023satlm, liu2023llm+, li2023guiding, huang2024rekep}. Related ideas have been explored in logical reasoning, planning, and embodied manipulation, where language models serve as translators into formal representations that can be checked or optimized by external solvers. Together, these works suggest that reliable spatial reasoning depends not only on stronger tools or richer geometry, but also on tighter control over the early query-to-requirement step before downstream execution.

\section{Evaluation Benchmark Details}
\label{app:benchmark}

\subsection{Benchmark Details}
\label{subsec:app_benchmarks}

\subsubsection{MMSI-Bench}
\label{subsec:app_mmsi}

MMSI-Bench~\cite{yang2025mmsi} evaluates multi-image spatial intelligence in real-world scenes. 
Each question is paired with multiple images captured from different viewpoints, requiring the model to integrate cross-view evidence rather than reason from a single observation. 
This benchmark is particularly relevant to our setting because many questions depend on resolving view-specific entities, camera viewpoints, and spatial relations that are only partially observable in each image. 
It contains four major categories:

\begin{itemize}
    \vspace{+0.5mm}
    \item \textbf{Positional Relationship.}
    This category tests whether a model can determine relative spatial relations among objects, cameras, and semantic regions across views. 
    Typical questions require identifying where one entity is located with respect to another after grounding both entities in the multi-view scene.

    \vspace{+0.5mm}
    \item \textbf{Attribute.}
    This category focuses on spatially related object attributes, including shape, size, length, and other geometry-dependent properties. 
    Unlike simple recognition tasks, these questions often require comparing or estimating physical properties from visual evidence.

    \vspace{+0.5mm}
    \item \textbf{Motion.}
    This category evaluates reasoning about object or camera movement. 
    The model must infer how a viewpoint or entity changes across images, which requires consistent alignment of observations over time or across views.

    \vspace{+0.5mm}
    \item \textbf{Multi-Step Reasoning.}
    This category combines multiple spatial operations into a longer reasoning chain. 
    Questions may require first grounding entities in different views, then resolving a reference frame, and finally computing or comparing a relation.
\end{itemize}

\subsubsection{ViewSpatial-Bench}
\label{subsec:app_viewspatial}

ViewSpatial-Bench~\cite{li2025viewspatial} is designed to evaluate viewpoint-sensitive spatial localization. 
Its questions emphasize that spatial terms such as left, right, front, and behind are not absolute scene properties, but depend on the viewpoint or agent-centered frame under which the relation is interpreted. 
This benchmark is useful for testing whether a model can distinguish externally anchored camera frames from human-centered or embodied perspectives.

\begin{itemize}
    \vspace{+0.5mm}
    \item \textbf{Camera-Based.}
    This split evaluates spatial reasoning from the perspective of a specified camera or image view. 
    The model must interpret directional relations according to the selected viewpoint, rather than relying on an implicit or default image frame.

    \vspace{+0.5mm}
    \item \textbf{Person-Based.}
    This split evaluates reasoning from a person-centered frame. 
    The queried relation must be interpreted with respect to the orientation or position of a human observer, making it especially sensitive to frame grounding and egocentric direction understanding.
\end{itemize}

\subsubsection{OMNI3D-Bench}
\label{subsec:app_omni3d}

OMNI3D-Bench~\cite{marsili2025visual} evaluates fine-grained 3D spatial reasoning from single images. 
Compared with multi-view benchmarks, this setting is more challenging in terms of depth ambiguity and metric estimation because the model must infer 3D layout from monocular visual evidence. 
The benchmark includes different answer formats, covering both categorical and numerical spatial reasoning.

\begin{itemize}
    \vspace{+0.5mm}
    \item \textbf{Counting.}
    This category tests whether a model can count spatially relevant entities under the constraints specified by the question. 
    It often requires grounding a candidate set and filtering objects according to visual or spatial conditions.

    \vspace{+0.5mm}
    \item \textbf{Float-MRA.}
    This category evaluates numerical 3D reasoning, such as estimating distances, sizes, or other continuous geometric quantities. 
    Following the benchmark protocol, performance is measured using mean relative accuracy rather than exact-match accuracy.

    \vspace{+0.5mm}
    \item \textbf{Yes/No.}
    This category asks binary spatial questions, requiring the model to verify whether a queried relation or geometric condition holds in the scene.

    \vspace{+0.5mm}
    \item \textbf{Multiple Choice.}
    This category provides a fixed set of candidate answers. 
    The model must select the option best supported by the inferred 3D relation, object layout, or metric property.
\end{itemize}

\section{Limitations}
PEPS reduces under-committed query compilation through a closed FESM schema, but this only shifts the failure mode rather than eliminating it. The library remains inherently bounded, and the conservative primitive-promotion rule means truly novel spatial concepts are under-served until sufficient evidence accumulates. The framework is also upper-bounded by its perception backbones: the failure analysis attributes most residual errors to orientation grounding, tool-workflow issues, and 3D reconstruction rather than to compilation itself, and the iteration-budget ablation confirms that the corrective loop cannot repair perception noise. Finally, the agent roles depend on a strong proprietary backbone, and the FESM schema is induced from a narrow slice of static-scene benchmarks, leaving embodied, dynamic, and out-of-distribution settings untested.

\section{Future Work}
Two directions follow naturally. First, making schema growth more automatic, for example through learned promotion criteria, would let the library evolve faster and extend more gracefully to new task families. Second, since the bottleneck has shifted from parsing to perception, the typed traces produced by PEPS could in turn be used to supervise or distill the perception backbones themselves, and to train smaller open-source agents that reduce dependence on proprietary APIs. Extending FESM with temporal and embodied primitives would broaden coverage to video, manipulation, and navigation, and the verifier's compilation-success signal could be developed into an explicit confidence score for selective prediction or human hand-off.
\newpage
\section{Qualitative Example}
\label{app:qualitativeexample}
\begin{figure}[h] % [htbp]
    \centering
    \vspace{-1em}
    \includegraphics[width=\textwidth]{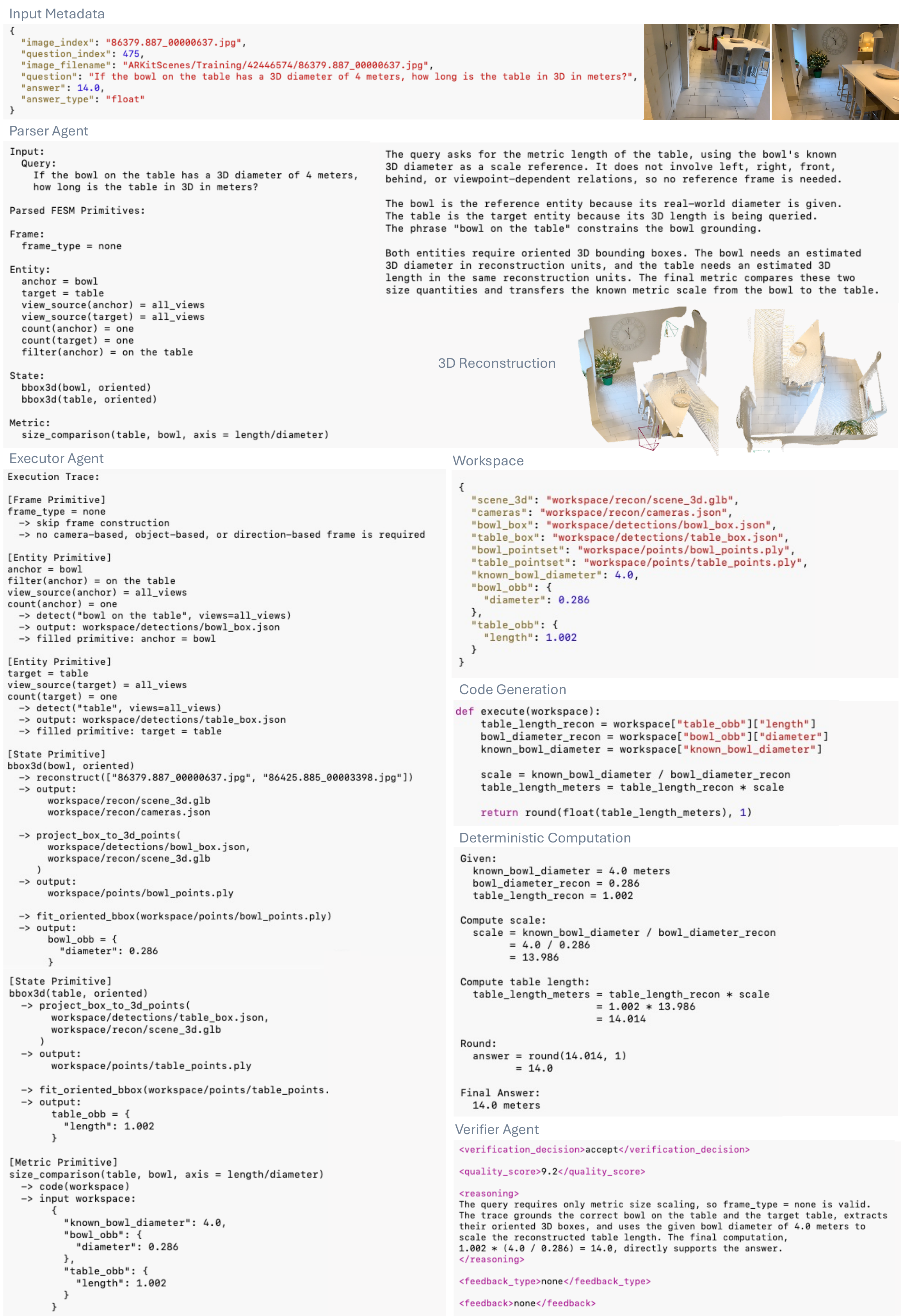} 
    % \vspace{-1.8em}
    \caption{Qualitative Example 1.} 
    \label{fig:qualitative example1}
    \vspace{-2em}
\end{figure}

\begin{figure}[h] % [htbp]
    \centering
    \vspace{-1em}
    \includegraphics[width=\textwidth]{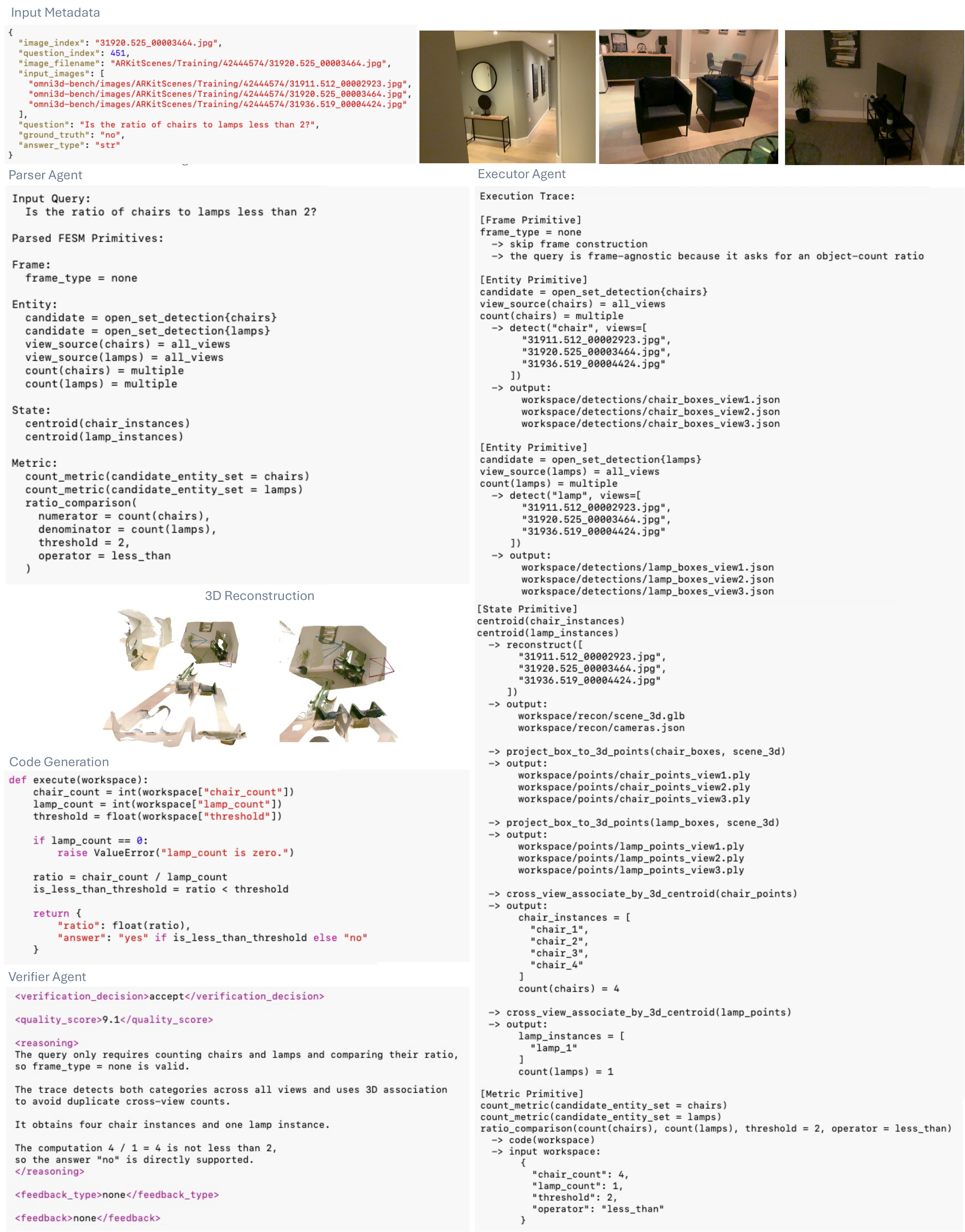} 
    % \vspace{-1.8em}
    \caption{Qualitative Example 2.} 
    \label{fig:qualitative example2}
    \vspace{-2em}
\end{figure}

\clearpage
\newpage
\section{Detailed FESM Primitive Schema}
\label{app:schema}

\begin{table*}[h]
\centering
\caption{
    \textbf{FESM Schema: Frame Primitives.}
    %
    % Our FESM (Frame--Entity--State--Metric) Schema defines a closed, typed space of \emph{spatial primitives} 
    % that a spatial reasoning query must be compiled into before any tool-based acquisition or deterministic 
    % computation takes place. 
    % %
    % This table lists the \textbf{Frame primitives}, which specify \emph{the coordinate frame under which the 
    % query should be interpreted}. In spatial reasoning, the same object pair can lead to different answers 
    % under different frames; therefore, resolving the reference frame is always the first step of query 
    % compilation. The remaining three groups of primitives (Entity, State, and Metric) are presented in the 
    % subsequent tables.
}
\begin{minipage}{0.99\linewidth}
\begin{tcolorbox}[colback=white]
\small

\textbf{[Frame]} \\
\textbf{Purpose}: These primitives determine which coordinate frame the query should be interpreted in. 
In spatial reasoning, the same object pair can lead to different answers under different frames. 
Therefore, before checking any entity state or relation metric, the system must first resolve the reference frame: 
\emph{whose viewpoint is used}, \emph{where the origin is}, and \emph{which direction counts as forward/up}. \\
\noindent\makebox[\linewidth]{\tikz[baseline]{\draw[dashed] (0,0) -- (\linewidth,0);}}

- \textbf{\textcolor{kleinblue2}{frame\_type}}: 
This primitive decides what kind of reference frame the query is asking for. 
It is the top-level choice that determines how the rest of the frame should be constructed, and it also 
uniquely determines the origin of the reference frame (i.e., the camera center for \texttt{camera\_based}, 
the anchor entity's representative point for \texttt{object\_based}, and the source entity's representative 
point for \texttt{direction\_based}). 
The valid values: \\
\hspace*{+3mm} - \texttt{[camera\_based]}: use the viewpoint of a specified image or camera. \\
\hspace*{+3mm} - \texttt{[object\_based]}: use the frame defined by an anchor object's pose, orientation, or interaction semantics. \\
\hspace*{+3mm} - \texttt{[direction\_based]}: use a frame whose forward direction is induced by an object-to-object direction or an object-to-cardinal alignment. \\
\hspace*{+3mm} - \texttt{[none]}: use when the query is frame-agnostic (e.g., counting, attribute, or scene-intrinsic queries) and no reference frame needs to be established. When selected, all other frame primitives are skipped. \\
Example: \texttt{\textcolor{green}{frame\_type=camera\_based}}, \texttt{\textcolor{green}{frame\_type=object\_based}}. \\

- \textbf{\textcolor{kleinblue2}{frame\_view}}: 
This primitive decides which image/viewpoint is the reference viewpoint when \texttt{frame\_type=camera\_based}. 
It tells the system whose camera frame should be used to interpret left/right/front/behind. 
Example: \texttt{\textcolor{green}{frame\_view=view\_1}}, \texttt{\textcolor{green}{frame\_view=image\_3}}. \\

- \textbf{\textcolor{kleinblue2}{frame\_anchor}}: 
This primitive decides which object defines the reference frame when \texttt{frame\_type=object\_based}. 
In other words, it identifies the object whose geometry, orientation, or interaction role provides the 
spatial anchor. 
Example: \texttt{\textcolor{green}{frame\_anchor=sofa}}, \texttt{\textcolor{green}{frame\_anchor=sink}}. \\

- \textbf{\textcolor{kleinblue2}{frame\_source}} / \textbf{\textcolor{kleinblue2}{frame\_target}}:
These two primitives decide which ordered object pair defines the reference direction. 
They are used when \texttt{frame\_type=direction\_based}, i.e., the frame is not taken from a camera or a single object, 
but from a directed relation between two objects. 
Conceptually, the forward direction is induced by the vector from source to target. 

\hspace*{+3mm} - Example: in ``oven is north of sink'', \texttt{\textcolor{green}{frame\_source=sink}} and 
\texttt{\textcolor{green}{frame\_target=oven}}. \\

- \textbf{\textcolor{kleinblue2}{frame\_forward}}:
This primitive decides what counts as the forward direction (\texttt{+Z\_frame}) in the reference frame. 
This is one of the most important frame primitives, because front/back/left/right all depend on how the 
positive forward axis is defined. The valid values include:\\
\hspace*{+3mm} - \texttt{[+Z\_{cam}]}: the camera view's front when \texttt{frame\_type=camera\_based}. \\
\hspace*{+3mm} - \texttt{[+Z\_obj]}: the anchor object's semantic front axis when \texttt{frame\_type=object\_based}. \\
\hspace*{+3mm} - \texttt{[V\_{source$\rightarrow$target}]}: set to \texttt{normalize(Centroid\_target - Centroid\_source)} when \texttt{frame\_type=direction\_based}. \\

- \textbf{\textcolor{kleinblue2}{frame\_down}}:
This primitive decides how the vertical axis (i.e., the positive down direction \texttt{+Y\_frame}, following 
the OpenCV convention) of the reference frame is determined. Unlike the forward direction, the vertical axis 
cannot always be inherited from a single anchor, because the reconstructed world frame is not guaranteed to be 
gravity-aligned. The valid values include:\\
\hspace*{+3mm} - \texttt{[+Y\_{cam}]}: follow the anchor camera's down axis when \texttt{frame\_type=camera\_based}. \\
\hspace*{+3mm} - \texttt{[+Y\_obj]}: follow the anchor object's semantic vertical axis when \texttt{frame\_type=object\_based}. \\
\hspace*{+3mm} - \texttt{[gravity]}: estimate the gravity direction from scene cues (e.g., detected ground plane, 
wall normals, or an auxiliary gravity-estimation tool). This value is required when \texttt{frame\_type=direction\_based} 
or when the query explicitly involves gravity-aligned concepts (e.g., \emph{above}, \emph{below}, \emph{top-down view}) 
and the anchor alone cannot provide a reliable vertical direction. \\

\end{tcolorbox}
\label{tab:frame_primitives}
\end{minipage}
\end{table*}

\begin{table*}[t]
\centering
\caption{
    \textbf{FESM Schema: Entity Primitives.}
}
\begin{minipage}{0.99\linewidth}
\begin{tcolorbox}[colback=white]
\small
\textbf{[Entity]} \\
\textbf{Purpose}: These primitives determine which scene entities the query is actually talking about. 
Before any geometric state or relation metric can be checked, the system must first ground the language 
in the query to concrete scene instances. 
These primitives resolve \emph{who is the reference entity}, \emph{who is the target}, 
\emph{where they should be grounded from}, and \emph{how ambiguous mentions should be narrowed down}. \\
\noindent\makebox[\linewidth]{\tikz[baseline]{\draw[dashed] (0,0) -- (\linewidth,0);}}

- \textbf{\textcolor{kleinblue2}{anchor}}:
This primitive decides which entity serves as the reference in the query. 
It identifies the entity relative to which another entity, direction, or relation will be interpreted. 
Its value is typically an entity category or a resolved entity mention extracted from the query. 
Note that the \texttt{anchor} is not necessarily the same as \texttt{frame\_anchor\_entity} in 
the Frame primitives: the former is the relational reference for the queried metric, while the latter 
defines the coordinate frame. A single query may involve both, and they may or may not coincide.\\
\hspace*{+3mm} - Example: \texttt{\textcolor{green}{anchor=sofa}}, \texttt{\textcolor{green}{anchor=sink}}. \\

- \textbf{\textcolor{kleinblue2}{target}}:
This primitive decides which entity is the actual queried target. 
It identifies the entity whose location, orientation, or spatial relation is being asked about. 
Its value typically corresponds to a single entity mention in the query. \\
\hspace*{+3mm} - Example: \texttt{\textcolor{green}{target=round dinner table}}, \texttt{\textcolor{green}{target=blue bin}}. \\

- \textbf{\textcolor{kleinblue2}{candidate}}:
This primitive decides which set of entities should be considered as possible answers. 
It is required when the query does not ask about one fixed target, but instead asks the system to search 
over a set of candidate entities (commonly triggered by queries such as ``which object \ldots'' or 
``what is \ldots''). 
To distinguish how the candidate set is constructed, this primitive takes two modes: \\
\hspace*{+3mm} - \texttt{[enumerated]}: the candidates are explicitly listed in the query itself (e.g., 
as multiple-choice options), and the set is taken directly from the query text. \\
\hspace*{+3mm} - \texttt{[open\_set\_detection]}: the candidates are specified only by a category prompt 
(e.g., ``all furniture'', ``any chair in the room''), and the set must be populated by open-vocabulary 
detection. \\
\hspace*{+3mm} - Example: 
\texttt{\textcolor{green}{candidate=[enumerated]\{table, painting, chair, desk\}}}, \\
\texttt{\textcolor{green}{candidate=[open\_set\_detection]\{all chairs\}}}. \\

- \textbf{\textcolor{kleinblue2}{view\_source}}:
This primitive decides from which input view(s) each entity should be grounded. 
It is essential for multi-view queries, where the same entity category may appear in multiple images 
and the query explicitly anchors an entity to a specific view (e.g., ``the chair in image~3''). 
Without this primitive, downstream detection cannot know which image to search in, and cross-view 
confusion becomes a major failure mode. 
The valid values include: \\
\hspace*{+3mm} - \texttt{[view\_id]}: a specific view index bound to the corresponding entity (e.g., 
\texttt{view\_1}, \texttt{image\_3}), used when the query ties the entity to one view. \\
\hspace*{+3mm} - \texttt{[all\_views]}: ground the entity across all available views, used when the query 
is view-agnostic or when cross-view association is needed. \\
\hspace*{+3mm} - Example: \texttt{\textcolor{green}{view\_source(target\_entity)=view\_3}}. \\

- \textbf{\textcolor{kleinblue2}{filter}}:
This primitive decides which scene region, local area, or contextual location the entity must belong to. 
It is needed when category, attribute, or view constraints alone are insufficient to uniquely identify 
the intended entity (e.g., ``the painting \emph{on the wall in the fitness area}''). 
Its value specifies a contextual localization constraint attached to a specific entity, to be consumed 
by downstream grounding as a disambiguation filter. \\
\hspace*{+3mm} - Example: 
\texttt{\textcolor{green}{filter(target\_entity)=\{on the wall, in the fitness area\}}}. \\

- \textbf{\textcolor{kleinblue2}{count}}:
This decides how many instances of a given entity the query expects to ground. 
It provides a cardinality constraint that the downstream grounding stage must enforce, and it is the 
primary trigger for ambiguity resolution: if the expected count does not match the actual number of 
detections, the system must invoke a disambiguation step (e.g., VLM-based re-selection using 
\texttt{entity\_region\_filter} or other cues) before proceeding. The valid values include: \\
\hspace*{+3mm} - \texttt{[one]}: exactly one instance is expected (e.g., ``the sofa''); if multiple are 
detected, disambiguation is mandatory. \\
\hspace*{+3mm} - \texttt{[multiple]}: more than one instance is expected (e.g., ``the chairs around the 
table''); the grounding should preserve all matching instances. \\
\hspace*{+3mm} - \texttt{[unspecified]}: the query does not impose any cardinality constraint, and any 
number of grounded instances is acceptable. \\
\hspace*{+3mm} - Example: \texttt{\textcolor{green}{count(target)=one}}. \\

\end{tcolorbox}
\label{tab:entity_primitives}
\end{minipage}
\end{table*}

\begin{table*}[t]
\centering
\caption{
    \textbf{FESM Schema: State Primitives.}
}
\begin{minipage}{0.99\linewidth}
\begin{tcolorbox}[colback=white]
\small
\textbf{[State]} \\
\textbf{Purpose}: These primitives determine which geometric states of the resolved cameras, entities, 
and scene must be acquired before any relation metric can be computed. 
The system should always follow the \emph{coarse-first, fine-on-demand} principle: retrieve the simplest 
state that is sufficient for the downstream metric, and only retrieve a finer state when the metric 
explicitly requires it (e.g., retrieve \texttt{centroid} instead of \texttt{pointset} 
unless surface-level geometry is needed; retrieve \texttt{orientation} instead of the full 
\texttt{3Dpose} unless full 6-DoF is required). \\
\noindent\makebox[\linewidth]{\tikz[baseline]{\draw[dashed] (0,0) -- (\linewidth,0);}}
- \textbf{\textcolor{kleinblue2}{camera\_pose(view)}}:
Returns the 3D pose of the specified camera view in the world frame, including camera position and 
orientation. 
This state is required for camera-based frames and for any view-dependent relation such as depth, 
visibility, or novel-view reasoning. It is typically provided directly by the output of the 
\texttt{reconstruct} tool. \\
\hspace*{+3mm} - Example: \texttt{\textcolor{green}{camera\_pose(view\_1)}}. \\

- \textbf{\textcolor{kleinblue2}{centroid(entity)}}:
Returns the representative 3D center of an entity as a single 3D point in the world frame. 
This is the coarsest and most frequently used location state. It should be the default choice 
whenever the downstream metric only depends on position (e.g., displacement, direction, centroid 
distance), without requiring the entity's full shape or orientation. \\
\hspace*{+3mm} - Example: \texttt{\textcolor{green}{centroid(chair)}}. \\

- \textbf{\textcolor{kleinblue2}{bbox3d(entity, mode)}}:
Returns a 3D bounding box describing the entity's occupied spatial extent. 
This state serves two purposes: it provides a compact proxy for entity size and coarse geometry, and 
it directly supports size-comparison and coarse-overlap queries. 
The \texttt{mode} argument specifies whether the box is axis-aligned to the world frame or aligned to 
the entity's own semantic axes, which yields different notions of ``size''. The valid values include:\\
\hspace*{+3mm} - \texttt{[axis\_aligned]}: an axis-aligned bounding box (AABB) in the world frame; 
suitable for rough spatial extent and coarse overlap. \\
\hspace*{+3mm} - \texttt{[oriented]}: an oriented bounding box (OBB) aligned with the entity's 
semantic axes; required when the query asks about width/length/height along the entity's own axes 
(e.g., ``how long is the sofa''). The oriented mode implicitly depends on \texttt{entity\_pose}. \\
\hspace*{+3mm} - Example: \texttt{\textcolor{green}{bbox3d(refrigerator, axis\_aligned)}}, 
\texttt{\textcolor{green}{bbox3d(sofa, oriented)}}. \\

- \textbf{\textcolor{kleinblue2}{pointset(entity)}}:
Returns the full 3D point set associated with the entity as an entity-level 3D point cloud in the 
world frame. 
This is the finest location-level state and should only be retrieved when a centroid or bounding box 
is insufficient---for example, when the downstream metric requires surface-to-surface distance, 
fine-grained occupancy, contact reasoning, or region-based overlap. \\
\hspace*{+3mm} - Example: \texttt{\textcolor{green}{pointset(table)}}. \\

- \textbf{\textcolor{kleinblue2}{3Dpose(entity)}}:
Returns the full 6-DoF pose of the entity in the world frame (position + rotation). 
This is the finest orientation-level state and should only be retrieved when the query requires the 
entity's complete local coordinate system, most notably when the entity itself defines an object-based 
reference frame (i.e., \texttt{frame\_type=object\_based} with \texttt{frame\_anchor\_entity=entity}), 
or when a local-frame transformation is needed. \\
\hspace*{+3mm} - Example: \texttt{\textcolor{green}{3Dpose(fireplace)}}. \\

- \textbf{\textcolor{kleinblue2}{orientation(entity)}}:
Returns a single 3D axis vector describing the entity's primary directional axis. 
This state is the lightweight alternative to \texttt{3Dpose} when only one direction is needed 
rather than the full 6-DoF. It automatically dispatches based on the entity's geometric type: \\
\hspace*{+3mm} - For volumetric entities (e.g., sofa, chair, appliance), it returns the entity's 
semantic front axis (i.e., the direction it is ``facing''). \\
\hspace*{+3mm} - For planar entities (e.g., painting, door, monitor, wall), it returns the outward 
surface normal, which plays the same role as the front axis for planar objects. \\
\hspace*{+3mm} - Retrieving both \texttt{orientation} and \texttt{3Dpose} for the 
same entity is redundant; the coarse-first principle requires the retriever to pick only one. \\
\hspace*{+3mm} - Example: \texttt{\textcolor{green}{orientation(sofa)}}, 
\texttt{\textcolor{green}{orientation(painting)}}. \\

\end{tcolorbox}
\label{tab:state_primitives}
\end{minipage}
\end{table*}

\begin{table*}[t]
\centering
\caption{
    \textbf{FESM Schema: Metric Primitives.}
}
\begin{minipage}{0.99\linewidth}
\begin{tcolorbox}[colback=white]
\scriptsize
% \small
\textbf{[Metric]} \\
\textbf{Purpose}: These primitives specify the geometric relation or quantity to be computed once 
the relevant frame, entities, and states are available. A query typically requires one top-level 
metric that directly answers the question, while other metrics may serve as intermediate quantities. \\
\noindent\makebox[\linewidth]{\tikz[baseline]{\draw[dashed] (0,0) -- (\linewidth,0);}}
- \textbf{\textcolor{kleinblue2}{relative\_displacement(entity\_a, entity\_b, frame)}}:
Returns the 3D displacement vector from \texttt{entity\_a} to \texttt{entity\_b} in the reference frame. 
Core metric for directional spatial relations. Computed from \texttt{entity\_centroid} of both entities. \\
\hspace*{+3mm} - Example: \texttt{\textcolor{green}{relative\_displacement(chair, blue\_bin, frame)}}. \\
- \textbf{\textcolor{kleinblue2}{planar\_displacement(entity\_a, entity\_b, frame)}}:
Returns the horizontal-plane displacement with the vertical component removed. Used for 
left/right/front/back and cardinal-direction reasoning. Derived from \texttt{relative\_displacement}.\\
\hspace*{+3mm} - Example: \texttt{\textcolor{green}{planar\_displacement(sofa, table, frame)}}. \\
- \textbf{\textcolor{kleinblue2}{vertical\_offset(entity\_a, entity\_b, frame)}}:
Returns the signed height difference along the frame's vertical axis. Required for above/below 
queries. Computed from \texttt{entity\_centroid} of both entities. \\
\hspace*{+3mm} - Example: \texttt{\textcolor{green}{vertical\_offset(counter, microwave, frame)}}. \\
- \textbf{\textcolor{kleinblue2}{euclidean\_distance(entity\_a, entity\_b)}}:
Returns the Euclidean distance between entity centroids. Frame-invariant. Used for near/far 
reasoning; when metric units are required, \texttt{estimate\_scale} is additionally invoked. Example: \texttt{\textcolor{green}{euclidean\_distance(chair, desk)}}. \\
- \textbf{\textcolor{kleinblue2}{surface\_distance(entity\_a, entity\_b)}}:
Returns the minimum surface-to-surface distance between two entities, more reliable than centroid 
distance when entity extent matters. Computed from \texttt{entity\_pointset} of both entities. Example: \texttt{\textcolor{green}{surface\_distance(sofa, coffee\_table)}}. \\
- \textbf{\textcolor{kleinblue2}{size\_comparison(entity\_a, entity\_b, axis)}}:
Returns the signed size difference along a specified axis (entity-local or world). Used for 
bigger/longer/taller queries. Computed from \texttt{entity\_bbox3d} of both entities. Example: \texttt{\textcolor{green}{size\_comparison(sofa, table, length)}}. \\
\noindent\makebox[\linewidth]{\tikz[baseline]{\draw[dashed] (0,0) -- (\linewidth,0);}}
- \textbf{\textcolor{kleinblue2}{orientation\_difference(axis\_1, axis\_2)}}:
Returns the angular difference between two orientation axes. Used for alignment analysis. Inputs 
typically come from \texttt{entity\_orientation\_axis} or a frame axis Example: 
\texttt{\textcolor{green}{orientation\_difference(entity\_orientation\_axis(chair), entity\_orientation\_axis(table))}}. \\
- \textbf{\textcolor{kleinblue2}{facing\_alignment(entity, target, frame)}}:
Returns a signed scalar in $[-1, 1]$ (cosine) between the entity's orientation axis and the direction 
toward \texttt{target} (another entity or a frame direction). The \texttt{code} tool discretizes it 
into categorical labels (facing toward / away / perpendicular). Computed from 
\texttt{entity\_orientation\_axis} and frame-transformed displacement. \\
\hspace*{+3mm} - Example: \texttt{\textcolor{green}{facing\_alignment(painting, north, frame)}}. \\
- \textbf{\textcolor{kleinblue2}{containment\_relation(entity\_inner, entity\_outer)}}:
Returns a Boolean and a containment ratio (fraction of inner points inside outer's bounding volume). 
Used for inside/within queries. Computed from \texttt{entity\_pointset(entity\_inner)} and 
\texttt{entity\_bbox3d(entity\_outer)}. \\
\hspace*{+3mm} - Example: \texttt{\textcolor{green}{containment\_relation(book, shelf\_compartment)}}. \\
- \textbf{\textcolor{kleinblue2}{between\_ratio(entity\_x, entity\_a, entity\_b, frame)}}:
Returns a scalar in $[0, 1]$ indicating how close \texttt{entity\_x} is to the midpoint of the segment 
between \texttt{entity\_a} and \texttt{entity\_b}, along with a perpendicular offset. Used for 
between-type queries. Computed from \texttt{entity\_centroid} of the three entities. Example: \texttt{\textcolor{green}{between\_ratio(chair, sofa, table, frame)}}. \\
- \textbf{\textcolor{kleinblue2}{count\_metric(candidate\_entity\_set)}}:
Returns the number of grounded instances in a candidate entity set. Used for ``how many'' queries. 
Computed from the cardinality of the grounded set after Entity-level disambiguation. Example: \texttt{\textcolor{green}{count\_metric(candidate\_entity\_set=\{chair\})}}. \\
\noindent\makebox[\linewidth]{\tikz[baseline]{\draw[dashed] (0,0) -- (\linewidth,0);}}
- \textbf{\textcolor{kleinblue2}{camera\_translation(view\_a, view\_b, frame)}}:
Returns the 3D translation vector of the camera from \texttt{view\_a} to \texttt{view\_b} in the 
reference frame. Used for ``in which direction did I move'' queries. Computed from 
\texttt{camera\_pose} of both views. \\
\hspace*{+3mm} - Example: \texttt{\textcolor{green}{camera\_translation(view\_1, view\_2, frame)}}. \\
- \textbf{\textcolor{kleinblue2}{camera\_rotation(view\_a, view\_b, frame)}}:
Returns the relative rotation of the camera from \texttt{view\_a} to \texttt{view\_b}, as a rotation 
matrix decoded by the \texttt{code} tool into rotation vector or Euler angles. Used for ``how did I 
turn'' queries. Computed from \texttt{camera\_pose} of both views. \\
\hspace*{+3mm} - Example: \texttt{\textcolor{green}{camera\_rotation(view\_1, view\_2, frame)}}. \\
\noindent\makebox[\linewidth]{\tikz[baseline]{\draw[dashed] (0,0) -- (\linewidth,0);}}
- \textbf{\textcolor{kleinblue2}{region\_object\_query(region\_spec, candidate\_entity\_set, frame)}}:
Returns the subset of entities in \texttt{candidate\_entity\_set} whose centroids fall inside a 3D 
region defined relative to an anchor entity or camera under the frame (e.g., ``left of \texttt{entity\_a}'', 
``in front of \texttt{view\_3}''). Used for ``what is to the left/right/front'' queries pervasive in 
multi-view spatial reasoning. Computed from \texttt{entity\_centroid} of each candidate and a 
region-membership test.\\
\hspace*{+3mm} - Example: \texttt{\textcolor{green}{region\_object\_query(left\_of(chair), 
candidate\_entity\_set=\{table, bin, sofa\}, frame)}}. 
\end{tcolorbox}
\label{tab:metric_primitives}
\end{minipage}
\end{table*}
\begin{table*}
\centering
\caption{
    \textbf{Prompt Used for Per-Question Primitive Decomposition.}
    This prompt is applied to each sampled question independently to decompose it into candidate spatial primitives across the four FESM layers. Here, \texttt{\{question\}} is the placeholder that will be replaced.
}
\begin{minipage}{0.99\linewidth}
\begin{tcolorbox} [colback=white]
    \small
    
    \textbf{[CORE MISSION]} \\
    You are an expert spatial reasoning analyst. Your sole mission is to analyze a user's spatial reasoning question and decompose it into the minimal set of information that a system must determine before the question can be solved deterministically. You should not attempt to solve the question. You should only identify what must be known. Ask yourself: \textit{``What are the irreducible pieces of information that any solver, regardless of its implementation, would need to commit to before computing the answer?''} \\
    \textbf{[OUTPUT FORMAT]} \\
    Your response \textbf{MUST} be a single, valid JSON object:
    \begin{alltt}\scriptsize
\{
  "reasoning": "A brief, step-by-step logical deduction that explains what 
                information the question depends on and why.",
  "primitives": \{
    "Frame": [\{"name": "...", "description": "...", "rationale": "..."\}],
    "Entity": [\{"name": "...", "description": "...", "rationale": "..."\}],
    "State": [\{"name": "...", "description": "...", "rationale": "..."\}],
    "Metric": [\{"name": "...", "description": "...", "rationale": "..."\}]
  \}
\}
\end{alltt}
    \vspace{-1.5mm} \textbf{[DECOMPOSITION FRAMEWORK]} \\
    Every spatial reasoning question implicitly requires the solver to commit to four categories of information. You must identify what the question demands in each category. If a category is not required by the question, return an empty list for that category. \vspace{+1mm} \\
    - \textbf{Frame}: What reference frame must be established to interpret the question? A reference frame is the coordinate system under which directional words such as \textit{left, right, front, behind, north, south} are grounded. Identify what defines the reference frame. For example, is it defined by a camera viewpoint, by an object's orientation, by a direction between two objects, or by none (if the question is frame-agnostic such as counting)? \vspace{+1mm} \\
    - \textbf{Entity}: Which concrete scene entities must the solver ground in order to answer the question? An entity is any object, region, or camera mentioned or implied by the question. Identify the anchor entity (the reference), the target entity (what is being asked about), and any disambiguating constraints (e.g., a region filter such as \textit{``on the wall''}, a view binding such as \textit{``in image 3''}, or a cardinality expectation such as \textit{``the sofa''} implying exactly one). \vspace{+1mm} \\
    - \textbf{State}: What geometric information about the entities must be acquired? Identify the minimum geometric state each entity requires. For example, does the question depend only on position (centroid), on full shape (point set), on size (bounding box or extent), or on orientation (pose or a single axis)? \vspace{+1mm} \\
    - \textbf{Metric}: What measurable quantity must ultimately be computed to answer the question? Identify the top-level relational computation. For example, is it a displacement vector, a distance, a vertical offset, an angular difference, a facing alignment, a containment test, a count, a camera motion, or a region-membership query? \vspace{+2mm} \\
    \textbf{[RULES]} \\
    - \textbf{Do not invent computational procedures}. You are identifying requirements, not solution steps. A good primitive is a piece of information the solver must know. A bad primitive is a step the solver must execute. \\
    - \textbf{Use concise, canonical names}. Name each primitive with a short, descriptive identifier in \texttt{snake\_case} (e.g., \texttt{anchor\_entity}, \texttt{entity\_orientation\_axis}, \texttt{planar\_displacement}). Avoid verbose or question-specific names. \\
    - \textbf{One primitive per distinct requirement}. If the question requires two different entities to be grounded, list both. If it requires two different geometric states (e.g., position of one entity and orientation of another), list both. \\
    - \textbf{Rationale must reference the question}. Each primitive's \texttt{rationale} must point to the specific phrase or semantic element in the question that demands it. \\
    - \textbf{No overlap across categories}. A reference frame definition belongs in Frame, not in Entity. A target object belongs in Entity, not in Metric. Keep each primitive in the category where its role is defined. \vspace{+2mm} \\
    %
%     \textbf{[EXAMPLE]} \\
%     %
%     Question: \textit{``From the viewpoint of image 3, what is to the left of the office chair?''} \\
%     Expected output (abbreviated):
%     \begin{alltt}\footnotesize
% \{
%   "reasoning": "The question asks for an object located relative to the chair
%                 under a camera-defined frame, so the solver must commit to the
%                 camera of image 3 as the reference viewpoint, ground the office
%                 chair, search over candidate entities, acquire their positions,
%                 and test region membership.",
%   "primitives": \{
%     "Frame": [
%       \{"name": "frame_type_camera_based", ...\},
%       \{"name": "frame_anchor_view", ...\}
%     ],
%     "Entity": [
%       \{"name": "anchor_entity", "description": "the office chair", ...\},
%       \{"name": "candidate_entity_set", ...\}
%     ],
%     "State": [
%       \{"name": "camera_pose", ...\},
%       \{"name": "entity_centroid", ...\}
%     ],
%     "Metric": [
%       \{"name": "region_object_query", ...\}
%     ]
%   \}
% \}
% \end{alltt}
%     %
%     \vspace{+1mm}
    \textbf{[QUESTION]} \\
    \{question\} \vspace{+2mm} \\
    Now, please analyze the above question and provide your response in the specified JSON format.
\end{tcolorbox}
\label{tab:app_prompt_per_question_decomposition}
\end{minipage}
\end{table*}

\begin{table*}
\centering
\caption{
    \textbf{Prompt Used for Cross-Question Schema Aggregation.}
    After per-question decomposition (Table~\ref{tab:app_prompt_per_question_decomposition}), this prompt aggregates the decomposition results across all sampled questions to induce the initial FESM schema. Here, \texttt{\{per\_question\_decompositions\}} is the placeholder for the concatenated JSON outputs from Stage A.
}

\begin{minipage}{0.99\linewidth}
\begin{tcolorbox} [colback=white]
    \scriptsize
    
    \textbf{[CORE MISSION]} \\
    You are an expert spatial reasoning analyst. You are given the per-question primitive decompositions of a large set of sampled spatial reasoning questions. Your sole mission is to aggregate these decompositions into a compact, non-redundant, typed schema that covers the spatial reasoning requirements observed across the questions. You should not add primitives that are not supported by the decompositions. You should not retain primitives that appear only as surface-level paraphrases of other primitives. Ask yourself: \textit{``What is the minimal set of typed spatial primitives under which every sampled question can be faithfully re-expressed?''} \\
    \textbf{[OUTPUT FORMAT]} \\
    Your response \textbf{MUST} be a single, valid JSON object: 
    \begin{alltt}
\{
  "reasoning": "A brief summary of how the per-question decompositions were
                grouped, merged, and organized into the final schema.",
  "schema": \{
    "Frame": [
      \{
        "name": "...",
        "description": "...",
        "valid_values": ["..."],
        "supporting_question_ids": [...],
        "coverage_count": N
      \}
    ],
    "Entity": [\{...\}],
    "State": [\{...\}],
    "Metric": [\{...\}]
  \},
  "discarded": [\{"name": "...", "reason": "merged into X" | "redundant with Y" | ...\}]
\}
\end{alltt}
    \vspace{-1.5mm} \textbf{[AGGREGATION PROCEDURE]} \\
    - \textbf{Group by semantic equivalence}: Primitives that appear under different names across questions but describe the same underlying requirement must be merged. For example, \texttt{anchor\_orientation}, \texttt{anchor\_facing\_axis}, and \texttt{reference\_front\_direction} all refer to the same concept and should be unified under one canonical primitive. \\
    - \textbf{Unify parameterization}: If two primitives differ only in their parameters or argument values (e.g., \texttt{centroid\_of\_chair} vs.\ \texttt{centroid\_of\_table}), merge them into a single parameterized primitive (e.g., \texttt{entity\_centroid(entity)}). \\
    - \textbf{Keep distinct what is truly distinct}: Two primitives that sometimes co-occur but can also appear independently across different questions must remain separate. Do not over-merge. \\
    - \textbf{Enumerate valid values where applicable}: If a primitive has a small closed set of values observed across the decompositions (e.g., \texttt{frame\_type} taking values \texttt{camera\_based}, \texttt{object\_based}, \texttt{direction\_based}), list them explicitly in \texttt{valid\_values}. \\
    - \textbf{Assign each primitive to exactly one FESM category}: Frame, Entity, State, or Metric. If a primitive appears to span two categories in some decompositions, inspect its core role and place it where it is defined. \\
    - \textbf{Record coverage}: For each primitive in the final schema, record the list of question IDs whose decomposition contributed to it, and the total count. This serves as evidence that the primitive is grounded in observed questions rather than speculated. \vspace{+2mm} \\
    \textbf{[CONSOLIDATION RULES]} \\
    - \textbf{Minimum support}: A primitive is included in the final schema only if it is supported by at least $\tau_{\text{support}}$ questions. Primitives with fewer supports are listed in the \texttt{discarded} field with reason \texttt{``insufficient support''}. \\
    - \textbf{No primitive without a computation path}: Each primitive in the final schema must be acquirable from the fixed geometric tool set available to downstream modules (3D reconstruction, detection, segmentation, object pose estimation, scale estimation, OCR, motion analysis, and code execution). If a candidate primitive cannot be grounded in these tools, list it in the \texttt{discarded} field with reason \texttt{``not tool-computable''}. \\
    - \textbf{Prefer coarse over fine when sufficient}: When two primitives express the same requirement at different granularities (e.g., \texttt{entity\_centroid} and \texttt{entity\_pointset}) and the coarser one is sufficient for most observed questions, retain both but mark the coarser one as the default. \\
    - \textbf{Canonical naming}: All final primitives must use short, descriptive \texttt{snake\_case} identifiers. Avoid verbose names or names that embed specific entity categories. \vspace{+2mm} \\
    \textbf{[PER-QUESTION DECOMPOSITIONS]} \\
    \{per\_question\_decompositions\} \vspace{+2mm} \\
    Now, please aggregate the above decompositions and provide the induced FESM schema in the specified JSON.
\end{tcolorbox}
\label{tab:app_prompt_schema_aggregation}
\end{minipage}
\end{table*}

\begin{table*}[t]
\centering
\caption{
    \textbf{Prompt Used for Retrieving FESM Requirements Library.}
    Here, \texttt{\{library\}} is the predefined FESM library, \texttt{\{examples\}} is the in-context examples,
    and \texttt{\{question\}} is the placeholder that will be replaced by the user's query with input images.
}
\begin{minipage}{0.99\linewidth}
\begin{tcolorbox}[colback=white]
\scriptsize

\textbf{[CORE MISSION]} \\
You are an expert spatial reasoning requirement analyst.
Your sole mission is to analyze the user's question together with the input images, and retrieve the minimal but sufficient set of required items from the predefined \textbf{Frame--Entity--State--Metric (FESM) library}.
Your goal is \textbf{NOT} to answer the question, \textbf{NOT} to perform geometric computation, \textbf{NOT} to decide tool calls, and \textbf{NOT} to verbalize the final answer.
Your task is only to determine: \textit{``What must the system know, ground, and acquire before this query can be solved deterministically?''}  \\

\textbf{[OUTPUT FORMAT]} \\
Your response \textbf{MUST} be a single, valid JSON object: 
\begin{alltt}\scriptsize
\{"reasoning": "A brief, step-by-step logical deduction explaining why the selected \\items are necessary and why no extra items are needed.",
  "requirements": \{
    "Frame": [
      "Exact library entries required to define the reference frame"
    ],
    "Entity": [
      "Exact library entries required to ground entities"
    ],
    "State": [
      "Exact library entries required to acquire geometric states"
    ],
    "Metric": [
      "Exact library entries required to compute the final relation"
    ]\}\}
\end{alltt}

\textbf{[RETRIEVAL PRINCIPLE]} \\
\textbf{1. Retrieve Requirements in Fixed Order} \\
You MUST retrieve requirements strictly in the following order: \textbf{Frame} $\rightarrow$ \textbf{Entity} $\rightarrow$ \textbf{State} $\rightarrow$ \textbf{Metric}. \\
- \textbf{Priority 1: Frame.} First determine under which reference frame the query should be interpreted. \\
- \textbf{Priority 2: Entity.} Then determine which concrete scene entities the query refers to. \\
- \textbf{Priority 3: State.} Then determine which geometric states must be acquired for those entities. \\
- \textbf{Priority 4: Metric.} Finally determine which measurable quantity should be computed to solve the query. \vspace{+0.5mm} \\
\textbf{2. Retrieve Frame Requirements} \\
To retrieve \textbf{Frame}, you MUST identify the element that gives final authority to spatial directions such as ``front'', ``behind'', ``left'', ``right'', or cardinal directions such as ``north''. You MUST retrieve only the entries necessary for the chosen frame type. Do not mix frame types unless the query explicitly requires it. \vspace{+0.5mm} \\
\textbf{3. Retrieve Entity Requirements} \\
To retrieve \textbf{Entity}, you MUST identify which scene entities are actually involved in the query. \\
- \textbf{A. Relation Anchor}: Retrieve the entity that serves as the spatial reference in the queried relation. \\
- \textbf{B. Target Entity}: Retrieve the entity whose position, orientation, or relation is being queried. \\
- \textbf{C. Candidate Set}: Retrieve a candidate set only when the query asks to search over multiple possible entities. \\
- \textbf{D. Disambiguation Constraints}: Retrieve region filters, attribute filters, or contextual constraints only when they are necessary to uniquely ground the intended entity. \\
- Do not assume that the frame anchor and the relation anchor are always the same. \vspace{+0.5mm} \\
\textbf{4. Retrieve State Requirements} \\
To retrieve \textbf{State}, you MUST identify which geometric states must be acquired before any computation. \\
- \textbf{A. Prefer Sufficient Primitive States}: Retrieve the simplest state that is sufficient for downstream computation. \\
- \textbf{B. Coarse States First}: Prefer coarse states such as centroid or 3D box when they are enough. \\
- \textbf{C. Fine States on Demand}: Retrieve pose, front axis, normal axis, point set, or camera pose only when the query explicitly depends on orientation, dense geometry, or viewpoint. \\
- Do not retrieve finer states if a simpler state is already sufficient. \vspace{+0.5mm} \\
\textbf{5. Retrieve Metric Requirements} \\
You MUST identify the final measurable quantity that directly solves the query once the required states are available. \\
% %
% - \textbf{A. Directness}: Retrieve the most direct metric for the queried relation. \\
% %
% - \textbf{B. Single-Metric Preference}: If one metric is sufficient, do not retrieve additional metrics. \\
% %
% - \textbf{C. Relation Matching}: The metric must match the relation type in the query, such as direction, distance, depth, orientation, containment, or between-ness. \vspace{+0.5mm} \\
% \textbf{6. Enforce Minimality and Sufficiency} \\
% %
% Your final retrieval must satisfy both of the following conditions: \\
% %
% - \textbf{A. Sufficiency}: The retrieved items must be enough for downstream value acquisition and deterministic computation. \\
% %
% - \textbf{B. Minimality}: The retrieved items must not contain unnecessary, duplicated, or trivially derivable entries. \\
% %
% - If a simpler item is sufficient, do not retrieve a more complex one. \\
% %
% - If a derived item can be obtained directly from an already selected item, do not retrieve it again unless explicitly required by the query.
%
\textbf{[LIBRARY]} \\
\{library\} \vspace{+2mm} \\
\textbf{[EXAMPLES]} \\
\{examples\} \vspace{+2mm}\\
\textbf{[QUESTION]} \\
\{question\} \vspace{+2mm}\\
Now, please analyze the above question and provide your response in the specified JSON format.

\end{tcolorbox}
\label{tab:prompt_fesm_retrieval}
\end{minipage}
\end{table*}

\begin{table*}[t]
\centering
\caption{
    \textbf{Prompt Used for Requirement-Guided Tool Orchestration.}
    Here, \texttt{\{api\_documents\}} is the documentation of available APIs,
    \texttt{\{requirements\}} is the retrieved requirement set,
    and \texttt{\{history\}} contains the user question, current workspace, previous plans, and execution results.
}
\begin{minipage}{0.99\linewidth}
\begin{tcolorbox}[colback=white]
\scriptsize

\textbf{[CORE MISSION]} \\
You are an expert spatial reasoning executor.
Your mission is to acquire the missing values required by the pre-retrieved \textbf{Frame--Entity--State--Metric (FESM)} requirements.
You are \textbf{not} allowed to freely redesign the solving procedure.
You must use the requirement set as the only target, inspect the current workspace, and generate the \textbf{immediate next tool calls} needed to fill the missing variables. \vspace{+2mm} \\
\textbf{[AVAILABLE APIS]} \\
\{api\_documents\} \vspace{+2mm} \\
\textbf{[REQUIREMENT SET]} \\
\{requirements\} \vspace{+2mm} \\
\textbf{[EXECUTION POLICY]} \\
\textbf{1. Follow the Requirement Set Strictly} \\
- The requirement set already specifies \textbf{what must be known}. You must not invent new goals beyond it. \\
- Your job is only to determine \textbf{how to acquire the still-missing values} needed by the requirement set. \\
- Always reuse variables already stored in the workspace before calling new tools. \vspace{+0.5mm} \\
\textbf{2. Acquire Required Values}: Based on the user's question and the pre-retrieved \textbf{Frame--Entity--State--Metric} requirement set, plan the necessary tool calls to acquire all missing values needed for deterministic computation. This process involves four dependency-coupled goals: \\
- \textbf{A. Frame Requirements}: First establish the reference frame. \\
    \hspace*{+3mm} - If the frame depends on a \textbf{camera}, call \texttt{reconstruct} to obtain reconstruction context and camera extrinsics. \\
    \hspace*{+3mm} - If the frame depends on an \textbf{object's orientation}, first ground that object, then call \texttt{predict\_obj\_pose}. \\
    \hspace*{+3mm} - If the frame depends on a \textbf{direction between two objects}, first ground both objects, then acquire their 3D representative points (e.g., centroids) using APIs such as \texttt{project\_box\_to\_3d\_points}. \\
%
% - \textbf{B. Entity Requirements}: Then ground the queried entities. \\
% %
%     \hspace*{+3mm} - Use detection or segmentation APIs to localize the required anchor, target, or candidate entities. \\
% %
%     \hspace*{+3mm} - If the query contains region or attribute constraints, use them only for disambiguation. \\
% %
% - \textbf{C. State Requirements}: Then acquire the required geometric states. \\
% %
%     \hspace*{+3mm} - For location-based queries, prefer simple states such as centroid or 3D box. \\
% %
%     \hspace*{+3mm} - For orientation-based queries, acquire pose, front axis, or normal axis. \\
% %
%     \hspace*{+3mm} - For view-dependent queries, acquire camera pose or rendered evidence if needed. \\
%
- \textbf{B. Entity Requirements}: Then ground the queried entities. \\
    \hspace*{+3mm} - Use \texttt{detect} to localize the required anchor entity, target entity, or candidate entity set in the image. \\
    \hspace*{+3mm} - If the query contains region, attribute, or text constraints, use them only for disambiguation. \\
    \hspace*{+3mm} - When a grounded entity is needed in 3D, call \texttt{project\_box\_to\_3d\_points} on its detected box to obtain its corresponding 3D points in the world frame; use \texttt{ocr} when text is part of the grounding clue. \\
- \textbf{C. State Requirements}: Then acquire the geometric states required by the retrieved requirements. \\
    \hspace*{+3mm} - For orientation-based states such as object pose or local axes, call \texttt{predict\_obj\_pose}. \\
    \hspace*{+3mm} - For metric-distance queries, call \texttt{estimate\_scale} before performing meter-level computation. \\
    \hspace*{+3mm} - For subtle view-change or small camera-motion queries, call \texttt{analyze\_motion}. \\
    \hspace*{+3mm} - Once all required state variables are available in the workspace, call \texttt{code} to compute the required metric. \\
- \textbf{D. Metric Requirements}: Finally prepare the inputs needed for the required metric. \\
    \hspace*{+3mm} - Only when all prerequisite variables are available, call \texttt{code} to compute the metric deterministically. \vspace{+0.5mm} \\
\textbf{4. Use Tools to Fill Variables, Not to Guess Answers} \\
- Tool calls must be used to obtain concrete variables required by the requirement set. \\
- Do not infer the final answer during this stage. \\
- Do not verbalize, judge, or conclude the queried relation before the required values are available. \vspace{+0.5mm} \\
%
% \textbf{5. Code Only When Ready} \\
% %
% - If the workspace already contains all variables, call \texttt{code} to perform deterministic computation. \\
% %
% - If computation has already been completed and the final result is available, call the final answer API. \\
% %
% - Otherwise, continue acquiring the next missing variables only. \vspace{+2mm} \\
%
\textbf{[OUTPUT FORMAT]} \\
Your response \textbf{MUST} be a single, valid JSON object: \vspace{-2.5mm}
\begin{alltt}
\{
  "analysis": "Briefly explain which required variables are still missing in the
               workspace, and why the proposed tool calls are the immediate next step.",
  "missing_requirements": [
    "List the unresolved requirement items that motivate the next action"
  ],
  "tool_calls": [
    \{
      "api": "API name",
      "args": \{...\},
      "output_variable": "A unique variable name to store the result in workspace"
    \}
  ]
\}
\end{alltt}
\textbf{[HISTORY]} \\
Here is the history so far: \{history\} \vspace{+2mm} \\
Please examine the requirement set and the current workspace, then generate the next-step response.
Your plan must include \textbf{only the immediate next action}, not the full solution.

\end{tcolorbox}
\label{tab:prompt_requirement_guided_tool}
\end{minipage}
\end{table*}

\begin{table*}[t]
\centering
\caption{
    \textbf{Prompt Used for Coder.}
    Here, \texttt{\{question\}} is the original user query, 
    \texttt{\{requirements\}} is the retrieved Frame--Entity--State--Metric requirement set,
    \texttt{\{workspace\_docs\}} documents the currently available variables in the workspace,
    and \texttt{\{knowledge\}} contains the fixed geometric formulas and rules relevant to the required metric.
}
\begin{minipage}{0.99\linewidth}
\begin{tcolorbox}[colback=white]
\small

\textbf{[CORE MISSION]} \\
You are an expert geometric computation programmer.
Your task is to write a single Python function that deterministically computes the result required by the retrieved \textbf{Frame--Entity--State--Metric} requirements, using only the variables already available in the workspace. \vspace{+2mm} \\
\textbf{[QUESTION]} \\
This is the original high-level query that provides the task context. \\
\{question\} \vspace{+2mm} \\
\textbf{[RETRIEVED REQUIREMENTS]} \\
The query has already been formalized into a \textbf{Frame--Entity--State--Metric} requirement set.
You must follow this requirement set strictly. \\
\{requirements\} \vspace{+2mm} \\
\textbf{[WORKSPACE VARIABLES]} \\
These are the variables that have already been acquired by previous tool calls.
Your code must use these variables as the only computational inputs. \\
\{workspace\_docs\} \vspace{+2mm} \\
\textbf{[FIXED GEOMETRIC KNOWLEDGE]} \\
These formulas and rules are valid and can be used directly in your implementation. \\
\{knowledge\} \vspace{+2mm} \\
\textbf{[AVAILABLE LIBRARIES]} \\
You may use \texttt{numpy}, \texttt{torch}, \texttt{scipy}, \texttt{math}, and other standard Python libraries. \vspace{+2mm} \\

\textbf{[CODING INSTRUCTIONS]} \\
\textbf{1. Compute Only from Available Variables} \\
- You must compute the result only from the variables documented in the workspace. \\
- Do not assume any missing variable, and do not invent any new input. \\
- If a required quantity must be derived, derive it from the provided variables. \vspace{+0.5mm} \\
\textbf{2. Respect the Retrieved Requirements} \\
- Follow the \textbf{Frame} requirements when interpreting directions such as left/right/north/south. \\
- Use the \textbf{Entity} requirements to ensure that computation is performed on the correct grounded entities. \\
- Use the \textbf{State} requirements to identify which geometric inputs are relevant. \\
- Use the \textbf{Metric} requirements to determine what must be computed as the final quantity. \vspace{+0.5mm} \\
\textbf{3. Perform Deterministic Computation} \\
- Your function must implement explicit geometric computation, not heuristic guessing. \\
- If coordinate transformation is required, transform the relevant variables into the required reference frame before judging the final relation. \\
- If the final result is categorical, compute the necessary geometric quantity first, then map it to the final label or option. \vspace{+0.5mm} \\
\textbf{4. Handle Multiple-Choice Questions Carefully} \\
- If the question is multiple-choice, your code must evaluate the given options systematically. \\
- Restrict the decision process to the provided options only. \\
- Do not introduce new answer candidates beyond the given choices. \vspace{+0.5mm} \\
\textbf{5. Output Only Code} \\
Your final output \textbf{MUST} be a single Python function in the following format: 
\begin{alltt}``` python
def execute({func_signature}):
    # Your code here
    ...
    return serializable_value
```\end{alltt}
The return value must be serializable.
Do not include any explanation, markdown, or extra text outside the function.

\end{tcolorbox}
\label{tab:prompt_coder_fesm}
\end{minipage}
\end{table*}

\renewcommand{\lstlistingname}{Table}
\lstdefinestyle{prompt}{
    % backgroundcolor=\color{gray!10},
    basicstyle=\footnotesize,
    breaklines=true,
    frame=single,
    frameround=tttt,
    framesep=10pt,
    % framerule=1.2pt,
    numbers=none,
    showstringspaces=false,
    keywordstyle=\color{blue!70!black}\bfseries,
    commentstyle=\color{green!50!black},
    stringstyle=\color{red!80!black},
    emphstyle=\color{orange!80!black}\bfseries,
    moredelim=[is][\color{red!70!black}]{\{}{\}},
    literate={<**>}{**}2,
    columns=fullflexible,
    escapeinside={(*@}{@*)},
    xleftmargin=0.4cm,
    xrightmargin=0.4cm,
}

\FloatBarrier
\refstepcounter{table}
\begin{lstlisting}[style=prompt, caption={Prompt for Verifier Candidate Evaluation}, label={prompt:verifier_judge}]
(*@\textbf{[CORE MISSION]}@*)
You are an expert verifier evaluating the quality of a program that solves a 3D spatial reasoning problem with tools and primitives. The candidate is a structured execution outcome consisting of: a predicted answer and corresponding execution trace. Your job is to: decide whether this candidate should be accepted or rejected; assign a quality score from 1.0 to 10.0; and, if rejected, return structured feedback that explains whether the failure comes from incomplete use of the existing primitive schema or from a primitive schema gap.
(*@\noindent\makebox[\linewidth]{\tikz[baseline]{\draw[dashed] (0,0) -- (\linewidth,0);}}@*)
(*@\textbf{[Question]}@*)
(*@{\{question\}}@*)

(*@\textbf{[Input Images]}@*)
(*@{\{Images\}}@*)

(*@\textbf{[Program to Evaluate]}@*)
(*@{\{\texttt{program\_code}\}}@*)

(*@\textbf{[Execution Trace]}@*)
- **Execution Status: (*@{\{success/failure/partial\}}@*)
- **Final Answer: (*@{\{answer\_text\}}@*)
- **Used APIs / Tools: (*@{\{used\_tools\}}@*)
- **Execution Error: (*@{\{execution\_error if present\}}@*)
(*@\noindent\makebox[\linewidth]{\tikz[baseline]{\draw[dashed] (0,0) -- (\linewidth,0);}}@*)
(*@\textbf{[EVALUATION FRAMEWORK]}@*)
 1. Query-Answer Consistency
Check whether the predicted answer actually addresses the user's question.
- Does the answer correspond to the queried relation, object, direction, or quantity?
- Is it semantically compatible with the query?
- If the question is multiple-choice, is the returned answer consistent with the provided options?
2. Visual Evidence Verification
Use the input images as grounding evidence throughout your evaluation. Check whether:
- the answer is visually plausible,
- the identified entities match the visible scene,
- the relevant spatial relation is compatible with the observed 3D structure,
- the intermediate results in the trace are visually consistent with the scene.
3. FESM Requirement Coverage
The retrieved primitives set defines what the system committed to knowing before solving the query.
You must check whether the execution trace sufficiently fulfills these commitments. Specifically verify:
- Frame: Was the reference frame fully grounded? If directional interpretation is required, is the frame anchor, forward axis, and vertical axis sufficiently established?
- Entity: Were the required entities correctly grounded and disambiguated? Were view constraints, region filters, or expected-count constraints respected?
- State: Were the required geometric states acquired at the appropriate granularity? Was a coarse state used where sufficient, and was a finer state retrieved when required?
- Metric: Was the final metric actually computable from the available instantiated states?
Pay close attention to any primitive marked as partial, failed, unresolved, or only weakly supported.
4. Failure Attribution
If the candidate should be rejected, you must identify which of the following two cases applies:
(*@\textbf{A. Missing-slot feedback}@*)
Use this when the current schema is already expressive enough, but the retrieved or instantiated primitive set is incomplete.
(*@\textbf{B. Primitive-gap feedback}@*)
Use this only when the current schema itself lacks the representational capacity needed by the query. Typical signs include:
- the query requires a type of frame, entity constraint, state, or metric that cannot be expressed by the current primitives,
- the trace fails not because a known primitive was omitted, but because no existing primitive could represent the need,
- repeated partial workarounds are visible, but the missing concept is not expressible within the current schema.
(*@\noindent\makebox[\linewidth]{\tikz[baseline]{\draw[dashed] (0,0) -- (\linewidth,0);}}@*)
(*@\textbf{[SCORING CRITERIA (1.0 - 10.0 Scale)]}@*)
1. Requirement Satisfaction
- How completely does the execution trace fulfill the retrieved FESM requirements?
- Are all necessary primitives properly instantiated?
2. Geometric Soundness
- Is the reasoning grounded in valid geometric evidence?
- Does the final computation follow from the instantiated states and metric?
3. Visual and Semantic Correctness
- Is the answer consistent with both the visual scene and the query semantics?
- Are grounded entities and relations plausible given the images?
4. Trace Reliability
- Does the trace show a clear and valid requirement-to-computation path?
- Are intermediate variables and tool outputs coherent and usable?
5. Diagnostic Confidence
- Can the success or failure be clearly attributed?
- If rejected, is the failure type identifiable as missing-slot or primitive-gap?
(*@\noindent\makebox[\linewidth]{\tikz[baseline]{\draw[dashed] (0,0) -- (\linewidth,0);}}@*)
(*@\textbf{[OUTPUT FORMAT]}@*)
You (*@\textbf{MUST}@*) provide your response in exactly this format:

\<verification_decision>accept|reject</verification_decision>
<quality_score>NUMBER</quality_score>
<reasoning>
[Detailed explanation covering:
- whether the answer is supported by the query, images, and execution trace,
- whether the retrieved FESM requirements are sufficiently fulfilled,
- whether the final computation is geometrically well-supported,
- what specific strengths or failures are visible in the trace,
- and why the candidate should be accepted or rejected.]
</reasoning>
<feedback_type>none|missing_slot|primitive_gap</feedback_type>
<feedback>
[If verification_decision=reject, provide a concise but precise structured feedback message.
Examples:
- "frame grounding incomplete; target entity grounded but anchor orientation missing"
- "metric requires planar projection before categorical left/right judgment"
- "current schema lacks event-defined orientation primitive"
If verification_decision=accept, output exactly: none]
</feedback>

\end{lstlisting}

\end{document}